# A Novel Fuzzy Approximate Reasoning Method Based on Extended Distance Measure in SISO Fuzzy System

I.M. SON, S.I. KWAK*, U.J. HAN, J.H. PAK, M. HAN, J.R. PYON, U.S. RYU

Faculty of Information Science, KIM IL SUNG University, Pyongyang, DPR KOREA
* Correspondent Author, si.kwak@ryongnamsan.edu.kp

**Abstract:** This paper presents an original method of fuzzy approximate reasoning that can open a new direction of research in the uncertainty inference of Artificial Intelligence(AI) and Computational Intelligence(CI). Fuzzy modus ponens (FMP) and fuzzy modus tollens(FMT) are two fundamental and basic models of general fuzzy approximate reasoning in various fuzzy systems. And the reductive property is one of the essential and important properties in the approximate reasoning theory and it is a lot of applications. This paper suggests a kind of extended distance measure (EDM) based approximate reasoning method in the single input single output(SISO) fuzzy system with discrete fuzzy set vectors of different dimensions. The EDM based fuzzy approximate reasoning method is consists of two part, i.e., FMP-EDM and FMT-EDM. The distance measure based fuzzy reasoning method that the dimension of the antecedent discrete fuzzy set is equal to one of the consequent discrete fuzzy set has already solved in other paper. In this paper discrete fuzzy set vectors of different dimensions mean that the dimension of the antecedent discrete fuzzy set differs from one of the consequent discrete fuzzy set in the SISO fuzzy system. That is, this paper is based on EDM. The experimental results highlight that the proposed approximate reasoning method is comparatively clear and effective with respect to the reductive property, and in accordance with human thinking than existing fuzzy reasoning methods.

## 1. Introduction

This paper presents an original method of fuzzy approximate reasoning that can open a new direction of research in the uncertainty inference of Artificial Intelligence (AI) and Computational Intelligence (CI). The fuzzy approximate reasoning is one of the important branch in the uncertainty inference. Fuzzy modus ponens (FMP) and fuzzy modus tollens (FMT) are two fundamental and basic models of general fuzzy approximate reasoning in various fuzzy system theory and applications. And the reductive property is one of the essential and important properties in the approximate reasoning theory and it's a lot of applications. [3,16,23,34,36,37] However many researched fuzzy approximate reasoning methods have some shortcomings [1,2,4,9,24]. For example, as presented in [22,23], the underlying semantic of CRI [38] is linguistically unclear, and its fuzzy reasoning result does completely not satisfy the reductive property. As pointed out in [10], shortcoming of the triple implication principle (TIP) method is that it cannot be applied in fuzzy control. [10,11,15,16,25,31,32,36,37,39] As presented in [22], the several fuzzy reasoning methods based on the fuzzy relation have a contradiction that they can be applied to the practical problem, for example fuzzy control, but do not satisfy the reductive property, vice versa. There are a lot of papers fuzzy reasoning methods based on similarity measure(SM) [7,9,27-29,34,35]. Their basic idea is to consider the similarity measure of the consequent $B^*(y)$ and the fuzzy reasoning conclusion $B^*(y)$ if the antecedent $A(x)$ is similar to the given premise $A^*(x)$ for FMP. This idea is right. But, as mentioned in [30], the fuzzy reasoning methods based on SM depend strongly on the similarity measure and the modification function, and do not completely satisfy the reductive property.

As presented in [32], due to many fuzzy reasoning methods based on SM do use nonlinear operators, the fuzzy sets of reasoning result are non-normal and non-convex ones. Therefore in fuzzy reasoning processing, linear operators must possibly be used. According to [2,8,18-21,30,33], a lot of fuzzy reasoning methods mathematically seem that they are all accompanied with a common shortcoming, that is, information loss. One of the reasons that do not satisfy the reductive property is to refer to losses of information occurred in reasoning processes. Therefore, information loss must possibly be reduced in fuzzy reasoning processing. And when studied fuzzy reasoning, the property that FMT is opposite to FMP must be considered. As shown in [7,19,20,23], the criterion function for checking of fuzzy reasoning results has only 2 values, i.e., '1' or 'O' for satisfaction of the reductive property, '0' or '×' for non-satisfaction of one. That is, this evaluation is too strict for the reductive property. Therefore criterion function for checking of fuzzy reasoning result must possibly be defined flexibly.

In the paper [40], to overcome shortcoming of the CRI and TIP, authors proposed a new Quintuple Implication Principle (QIP) for fuzzy reasoning, which draws the approximate reasoning conclusion $B^*(y)$ of the consequent $B(y)$ (resp. $A^*(x)$ of $A(x)$) as the formula which is best supported by $A(x) \to B(y)$, $A^*(x) \to A(x)$, and $A^*(x)$ (resp. $A(x) \to B(y)$, $B(y) \to B^*(y)$, and $B^*(y)$), for FMP(resp. FMT). The proposed QIP was illustratively compared with the CRI and TIP solutions, which is much closer to the proposition that for example "x is small" and, therefore, in accordance with human thinking than CRI and TIP.

In order to overcome the previous various shortcomings presented in [2,7,8,10,17-23,27,28,30,36, and 40], a fuzzy reasoning method based on new idea has developed in the paper [12,13], which is without some losses of information due to use liner operators, and with smooth evaluation for the reductive property. That is, in the paper [12,13], authors proposed a new fuzzy reasoning method based on Turksen and Zhong's Euclidian DM [5,27,28], so called DMM, which consists of both DM for FMP and DM for FMT, for short, FMP-DM, and FMT-DM. The idea of the paper [12,13] is based on the paper [14].

In this paper we try to suggest a kind of distance measure based fuzzy approximate reasoning method in the single input single output (SISO) fuzzy system with discrete fuzzy set vectors of different dimensions. For this method, we call it an extended distance measure based fuzzy approximate reasoning method, for short, EDM method. That is, EDM method is consisted of two part, i.e., FMP-EDM, and FMT-EDM. As mentioned above, EDM based fuzzy reasoning method that the dimension of the antecedent discrete fuzzy set is equal to one of the consequent discrete fuzzy set has already solved in the paper [12,13]. In this paper, discrete fuzzy set vectors of different dimensions mean that the dimension of the antecedent discrete fuzzy set differs from one of the consequent discrete fuzzy set in the SISO fuzzy system. We compare the reductive properties for 5 fuzzy reasoning methods with respect to FMP and FMT, which are CRI, TIP, QIP, AARS, and an our proposed novel EDM method. The experimental results highlight that the proposed approximate reasoning method is comparatively clear and effective, and in accordance with human thinking than other reasoning methods.

The rest parts of this paper are organized as follows. In section 2, we discuss related works of distance measure and the method of the fuzzy modus ponens and fuzzy modus tollens. And we show about the evaluation based on the criterion function of fuzzy reasoning methods. In section 3, a novel fuzzy approximate reasoning method for FMP and FMT are proposed, i.e., FMP-EDM, and FMT-EDM, respectively, and then those reductive properties is proved, examples are showed for the proposed method. In section 4, the reductive property of Zadeh's CRI based method [38], Wang's TIP based method [32], Zhou et al.'s the QIP based method [40], Turksen et al.'s AARS method [27,28], and our proposed EDM method for FMP and FMT, is compared illustratively with respect to the reductive property. In section 5, we describe the conclusion of this paper.

## 2. Related Works

According to the paper [5,6,26,31], the general DM is described as follows. Let $F_0(R)$ be all continuous fuzzy subsets of $R$ whose $\alpha$-cuts are always bounded intervals. These will be called fuzzy numbers and are the fuzzy sets most widely used in practical applications. We need to be able to compute the distance between any fuzzy set $A$ and $B$ in $F_0(R)$. We know how to find the distance between two real numbers $x, y$. The distance is $|x-y| = DM(x,\ y)$. We also know how to find the distance between two points in $R^2$. The function $DM(x,\ y)$ used to compute distance is called a distance measure (DM). The basic properties of DM, i.e., $DM(x,\ y)$ for every $x, y$ in real space $R$ are:

- $DM(x,\ y) \geq 0$; i.e., distance is not negative;
- $DM(x,\ y) = DM(y,\ x)$; i.e., distance is symmetric;
- $DM(x,\ y) = 0$; if and only if $x = y$, i.e., we get zero distance only when $x = y$, i.e., $DM(x,\ y) = 0 \Leftrightarrow x = y$;
- $DM(x,\ y) \leq DM(x,\ z) + DM(z,\ y)$; i.e., it is shorter to go directly from *x* to *y* instead of first going to intermediate point *z*.

Generally known fuzzy reasoning methods are FMP and FMT in the fuzzy system with 1 input 1 output 1 rule.

General form of the fuzzy modus ponens presented in [7] is expressed as follows.

$$\text{Rule; } if\ x\ is\ A\ \ then\ y\ is\ B, \quad \text{Premise}: x\ is\ A^*, \quad \text{Conclusion}: y\ is\ B^* \tag{1}$$

General form of FMT presented in the paper [7] is described as follows.

$$\text{Rule; } if\ x\ is\ A\ \ then\ y\ is\ B, \quad \text{Premise}: y\ is\ B^*, \quad \text{Conclusion}: x\ is\ A^* \tag{2}$$

, where $A^* \in F(X),\ A \in F(X)$ are fuzzy sets defined in the universe of discourse $X$, $B^* \in F(Y),\ B \in F(Y)$ are fuzzy sets defined in the universe of discourse $Y$. In the fuzzy system with 1 input 1output n rules, we rewrite the definition for reductive property of fuzzy inference method presented in [7]. According to [7], the formula (2) can be written as follows, because FMT is opposite to FMP.

$$\text{Rule; } if\ y\ is\ \overline{B}\ \ then\ x\ is\ \overline{A}, \quad \text{Premise}: y\ is\ \overline{B}, \quad \text{Conclusion}: x\ is\ \overline{A} \tag{3}$$

, where $\overline{A} = 1 - A$, $\overline{B} = 1 - B$. The most general forms of the CRI solutions of FMP and FMT are as follows.

$$\text{(FMP-CRI)} \quad B^*(y) = \bigvee_{x \in U} (A^*(x) \otimes (A(x) \to B(y))) \tag{4}$$

$$\text{(FMT-CRI)} \quad A^*(x) = \bigvee_{y \in V} (B^*(y)) \otimes (A(x) \to B(y))) \tag{5}$$

Suppose $\otimes$ is a left continuous t-norm and $\to$ its residual. Then the TIP solution of FMP and FMT are expressed

as follows:

$$\text{(FMP-TIP)} \quad B^*(y) = \bigvee_{x \in U} (A^*(x) \otimes (A(x) \to B(y))) \tag{6}$$

$$\text{(FMT-TIP)} \quad A^*(x) = \bigwedge_{y \in V} ((A(x) \to B(y)) \to B^*(y))) \tag{7}$$

The QIP solution of FMP and FMT presented in [40] is described as follows:

$$\text{(FMP-QIP)} \quad B^*(y) = \bigvee_{x \in U} (A^*(x) \otimes (A^*(x) \to A(x)) \otimes (A(x) \to B(y))) \tag{8}$$

$$\text{(FMT-QIP)} \quad A^*(x) = \bigvee_{y \in V} (A \ (x) \otimes (A \ (x) \to B(y)) \otimes (B(y) \to B^*(y))) \tag{9}$$

Unlike CRI [38], in [28], a similarity-based fuzzy reasoning method, i.e., Turksen and Zhong's Approximate Analogical Reasoning Schema (AARS) was proposed. The AARS modifies the consequent based on the similarity (closeness) between the given premise $A^*$ and the antecedent $A$. If the rule is fired, then the consequent is modified by a modification function which could appear in one of the two forms for FMP and FMT i.e. *more or less form* and, *fuzzy membership value reduction form*, for short, *reduction form*, according to [28], respectively:

$$\text{(FMP-AARS-}\textit{more or less form}\text{)} \quad B^* = \min\{1, \ B/S_{AARS}\} \tag{10}$$

$$\text{(FMT-AARS-}\textit{more or less form}\text{)} \quad A^* = \min\{1, \ A/S_{AARS}\} \tag{11}$$

$$\text{(FMP-AARS-}\textit{reduction form}\text{)} \quad B^* = B \times S_{AARS} \tag{12}$$

$$\text{(FMT-AARS-}\textit{reduction form}\text{)} \quad A^* = A \times S_{AARS} \tag{13}$$

According to [28], one of distance measures (DM) for FMP, for FMT are as follows, respectively.

$$DM = D_2(A^*, A) = \left[ \sum_{i=1}^{n} [\mu_{A^*}(x_i) - \mu_A(x_i)]^2 / n \right]^{1/2}, \text{ for FMP} \tag{14}$$

$$DM = D_2(B^*, B) = \left[ \sum_{i=1}^{n} [\mu_{B^*}(y_i) - \mu_B(y_i)]^2 / n \right]^{1/2}, \text{ for FMT} \tag{15}$$

The similarity based on DM is then defined as follows.

$$S_{AARS} = (1 + DM)^{-1} \tag{16}$$

The reductive property is one of the essential properties in the applications of the fuzzy approximate reasoning [7,12,13,23]. The reductive property criteria for FMP and FMT is shown in Table 1.

**Table 1.** Reductive property criterion for FMP and FMT based on [7,12,13,23]

| FMP | *if x is A then y is B* | | |
|---|---|---|---|
| | $x$ is $A^*$ (premise) | $y$ is $B^*$ (conclusion ) | Reductive property criterion function $RPCF_{FMP}$ of $y$ is $B^*$, (%) |
| Case 1 | $A^* = A$ | $B^* = B$ | $RPCF_{FMP} = (1 - \sum_{k=1}^{\Theta} \lvert b^*_{kl} - b_k \rvert / \Theta) \times 100$ |
| Case 2 | $A^* = A^2$ | $B^* = B^2$ *or* $B$ | $RPCF_{FMP} = (1 - \sum_{k=1}^{\Theta} \lvert b^*_{kl} - b_k^2 \rvert / \Theta) \times 100$ or $(1 - \sum_{k=1}^{\Theta} \lvert b^*_{kl} - b_k \rvert / \Theta) \times 100$ |
| Case 3 | $A^* = A^{1/2}$ | $B^* = B^{1/2}$ *or* $B$ | $RPCF_{FMP} = (1 - \sum_{k=1}^{\Theta} \lvert b^*_{kl} - b_k^{\frac{1}{2}} \rvert / \Theta) \times 100$ or $(1 - \sum_{k=1}^{\Theta} \lvert b^*_{kl} - b_k \rvert / \Theta) \times 100$ |
| Case 4 | $A^* = 1 - A$ | $B^* = 1 - B$ | $RPCF_{FMP} = (1 - \sum_{k=1}^{\Theta} \lvert b^*_{kl}(1 - b_k) \rvert / \Theta) \times 100$ |
| Case 5 | $A^* = s.t. A$ | $B^* = s.t. B$ | $RPCF_{FMP} = (1 - \sum_{k=1}^{\Theta} \lvert b^*_{kl} - s.t. b_k \rvert / \Theta) \times 100$ |
| FMT | *if y is* $\overline{B}$ *then x is* $\overline{A}$ | | |
| | $y$ is $B^*$ (premise) | $x$ is $A^*$ (conclusion) | Reductive property criterion function $RPCF_{FMT}$ of $x$ is $A^*$, (%) |
| Case 6 | $B^* = 1 - B$ | $A^* = 1 - A$ | $RPCF_{FMT} = (1 - \sum_{k=1}^{\Theta} \lvert a^*_{kl} - (1 - a_k) \rvert / \Theta) \times 100$ |
| Case 7 | $B^* = 1 - B^2$ | $A^* = 1 - A^2$ *or* $1 - A$ | $RPCF_{FMT} = (1 - \sum_{k=1}^{\Theta} \lvert a^*_{kl} - (1 - a_k)^2 \rvert / \Theta) \times 100$ or $(1 - \sum_{k=1}^{\Theta} \lvert a^*_{kl} - (1 - a_k) \rvert / \Theta) \times 100$ |
| Case 8 | $B^* = 1 - B^{1/2}$ | $A^* = 1 - A^{1/2}$ *or* $1 - A$ | $RPCF_{FMT} = (1 - \sum_{k=1}^{\Theta} \lvert a^*_{kl} - (1 - a_k)^{1/2} \rvert / \Theta) \times 100$ or $(1 - \sum_{k=1}^{\Theta} \lvert a^*_{kl} - (1 - a_k) \rvert / \Theta) \times 100$ |
| Case 9 | $B^* = B$ | $A^* = A$ | $RPCF_{FMT} = (1 - \sum_{k=1}^{\Theta} \lvert a^*_{kl} - a_k \rvert / \Theta) \times 100$ |
| Case 10 | $B^* = s.t. B$ | $A^* = s.t. A$ | $RPCF_{FMT} = (1 - \sum_{k=1}^{\Theta} \lvert a^*_{kl} - s.t. a_k \rvert / \Theta) \times 100$ |

In the Table 1, the indexes are $\Theta = \{u \; or \; v \; or \; u \cdot v\}$, $l = 1, 2, \cdots$. And Class 1 contains Case 1, 2, 3, and 4, for FMP, and Case 6, 7, 8, and 9, for FMT, and Class 2 contains Case 1, 2, 3, and 5, for FMP, and Case 6, 7, 8, and 10, for FMT, respectively. And s.t. is an abbreviated word 《slightly tilted of》 according to [7]. And reductive property criterion function $RPCF_{FMP}$ of $y \; is \; B^*$, (%) is presented in [12,13].

## 3. A Novel Fuzzy Approximate Reasoning Method

In this section, we propose a kind of distance measure based fuzzy approximate reasoning method in SISO fuzzy system with discrete fuzzy set vectors of different dimensions. This method is called an extended distance measure(EDM) based fuzzy approximate reasoning method. Our proposed EDM based fuzzy approximate reasoning method is consisted of two part, i.e., FMP-EDM, and FMT-EDM. The fuzzy reasoning method that the dimension of the antecedent discrete fuzzy set is equal to one of the consequent discrete fuzzy set has already solved in the paper [12,13]. In this section, the discrete fuzzy set vectors of different dimensions mean that the dimension of the antecedent discrete fuzzy set differs from one of the consequent discrete fuzzy set in the SISO fuzzy system.

### 3.1. A Novel Fuzzy Approximate Reasoning Method For FMP-EDM

In this subsection, about the SISO fuzzy system with discrete fuzzy set vectors of different dimensions, we propose a novel fuzzy modus ponens based on extended distance measure method, for short, FMP-EDM method in the case that differs from dimensions between the antecedent and consequent, i.e., in the case of index $u \neq v$ when element number of the antecedent is $u$ and element number of the consequent is $v$.

Let us promise several concepts of FMP-EDM for an approximate reasoning in SISO fuzzy system with discrete fuzzy set vectors of different dimensions.

- Let $X$ be universe of discourse and $A$ $(A \subseteq X)$ be membership function of the antecedent fuzzy set of $X$, then $A = [a_i]_{u\times 1}, i = \overline{1, u}$ is called a $u \times 1$ dimension antecedent row vector.
- Let $Y$ be universe of discourse and $B$ $(B \subseteq Y)$ be membership function of the consequent fuzzy set of $Y$, then $B = [b_j]_{v\times 1}, j = \overline{1, v}$ is called a $v \times 1$ dimension consequent row vector.
- Let $X$ be universe of discourse and $A^* = \{A_l^*\}$ $(A_l^* \subseteq X, l = 1, 2, \cdots)$ be membership function of the given premise fuzzy set of $X$, then $A_l^* = [a_{il}^*]_{u\times 1}, i = \overline{1, u}, l = 1, 2, \cdots$ is called a $u \times 1$ dimension premise row vector.
- Let $Y$ be universe of discourse and $B^* = \{B_l^*\}$ $(B_l^* \subseteq Y, l = 1, 2, \cdots)$ be membership function of the conclusion fuzzy set of $Y$, then $B_l^* = [b_{jl}^*]_{v\times 1}, j = \overline{1, v}, l = 1, 2, \cdots$ is called a $v \times 1$ dimension conclusion row vector.
- Among fuzzy sets $A, A^*, B, \text{and} \, B^*$, their elements, i.e., values of membership functions $a_i \in A, a_{il}^* \in A_l^*, i = \overline{1, u}, b_j \in B, \text{and} \, b_{jl}^* \in B_l^*, j = \overline{1, v}$ are all the values of closed interval [0, 1] that characterizes the antecedent, the given premise, the consequent and new conclusion obtained by approximate reasoning, respectively.
- For SISO fuzzy system with discrete fuzzy set vectors of different dimensions, the antecedent, the given premise, the consequent and a new conclusion fuzzy row vectors can be expressed as follows; $A = [a_1, \cdots, a_i, \cdots, a_u]$, $A^* = \{A_l^*\}, A_l^* = [a_{1l}^*, \cdots, a_{il}^*, \cdots, a_{ul}^*], l = 1, 2, \cdots$, $B = [b_1, \cdots, b_j, \cdots, b_v]$, and $B^* = \{B_l^*\}, B_l^* = [b_{1l}^*, \cdots, b_{jl}^*, \cdots, b_{vl}^*], l = 1, 2, \cdots$, respectively.

A novel approximate reasoning method of FMP for SISO fuzzy system with discrete fuzzy set vectors of different dimensions is described as following steps, which is so called FMP-EDM.

**Step 1**; Compute the extended fuzzy row vectors.

Let $u, v, n$ be all real integer index, then the extended fuzzy sets of the antecedent $A$, the given premise $A^*$ and consequent $B$, i.e., the extended fuzzy vectors $\widetilde{A}$, $\widetilde{A}^* = \{\widetilde{A}_l^*\}, \widetilde{A}_l^* = [\widetilde{a}_{1l}^*, \cdots, \widetilde{a}_{il}^*, \cdots, \widetilde{a}_{ul}^*], l = 1, 2, \cdots$, and $\widetilde{B}$ are calculated as following 3 conditions, respectively;

- Condition 1; If index $u > v$, i.e., $u = n \cdot v$ then the extended every vectors $\widetilde{A}$, $\widetilde{A}^*$, and $\widetilde{B}$ are calculated as follows, respectively.

$$\widetilde{A} = A = \left[\frac{a_1}{x_1}, \frac{a_2}{x_2}, \cdots, \frac{a_i}{x_i}, \cdots, \frac{a_{u-1}}{x_{u-1}}, \frac{a_u}{x_u}\right], \text{ or simply, } \widetilde{A} = A = [a_1, a_2, \cdots, a_{u-1}, a_u] \tag{17}$$

$$\widetilde{A}_l^* = A_l^* = \left[\frac{a_{1l}^*}{x_1}, \frac{a_{2l}^*}{x_2}, \cdots, \frac{a_{il}^*}{x_i}, \cdots, \frac{a_{u-1,l}^*}{x_{u-1}}, \frac{a_{u,l}^*}{x_u}\right], \text{ or simply, } \widetilde{A}_l^* = A_l^* = [a_{1l}^*, a_{2l}^*, \cdots, a_{u-1,l}^*, a_{u,l}^*] \tag{18}$$

$$\widetilde{B} = \left[\frac{b_1}{y_1}, \frac{b_2}{y_2}, \cdots, \frac{b_p}{y_p}, \cdots, \frac{b_{n\cdot v-1}}{y_{n\cdot v-1}}, \frac{b_{n\cdot v}}{y_{n\cdot v}}\right], (\neq B, p = \overline{1, n\cdot v}), \text{ or simply, } \widetilde{B} = [b_1, b_2, \cdots, b_{n\cdot v-1}, b_{n\cdot v}] \tag{19}$$

- Condition 2; If index $u < v$, i.e., $v = n \cdot u$ then the extended every vectors $\widetilde{A}$, $\widetilde{A}^*$, and $\widetilde{B}$ are calculated as follows, respectively.

$$\widetilde{A} = \left[\frac{a_1}{x_1}, \frac{a_2}{x_2} \cdots, \frac{a_q}{x_q}, \cdots, \frac{a_{n\cdot u-1}}{x_{n\cdot u-1}}, \frac{a_{n\cdot u}}{x_{n\cdot u}}\right], (\neq A, q = \overline{1, n\cdot u}), \text{ or simply, } \widetilde{A} = \left[a_1, \cdots, a_q, \cdots, a_{n\cdot u}\right] \quad (20)$$

$$\widetilde{A}_l^* = \left[\frac{a_{1l}^*}{x_1}, \frac{a_{2l}^*}{x_2}, \cdots, \frac{a_{ql}^*}{x_q}, \cdots, \frac{a_{(n\cdot u-1),l}^*}{x_{n\cdot u-1}}, \frac{a_{(n\cdot u),l}^*}{x_{n\cdot u}}\right], (\neq A, q = \overline{1, n\cdot u}), l = 1, 2, \cdots, \text{ or simply,}$$

$$\widetilde{A}_l^* = \left[a_{1l}^*, \cdots, a_{il}^*, \cdots, a_{(n\cdot u),l}^*\right] \quad (21)$$

$$\widetilde{B} = B = \left[\frac{b_1}{y_1}, \frac{b_2}{y_2}, \cdots, \frac{b_j}{y_j}, \cdots, \frac{b_{v-1}}{y_{v-1}}, \frac{b_v}{y_v}\right], \text{ or simply, } \widetilde{B} = B = \left[b_1, \cdots, b_j, \cdots, b_v\right] \quad (22)$$

- Condition 3; If index $u \neq v$, i.e., $u \neq n \cdot v$ or $v \neq n \cdot u$ then the extended every vectors $\widetilde{A}$, $\widetilde{A}^*$, and $\widetilde{B}$ are calculated as follows, respectively.

$$\widetilde{A} = \left[\frac{a_1}{x_1}, \frac{a_2}{x_2}, \cdots, \frac{a_r}{x_r}, \cdots, \frac{a_{u\cdot v-1}}{x_{u\cdot v-1}}, \frac{a_{u\cdot v}}{x_{u\cdot v}}\right], (\neq A, r = \overline{1, u\cdot v}), \text{ or simply, } \widetilde{A} = \left[a_1, \cdots, a_r, \cdots, a_{u\cdot v}\right] \quad (23)$$

$$\widetilde{A}_l^* = \left[\frac{a_{1l}^*}{x_1}, \frac{a_{2l}^*}{x_2}, \cdots, \frac{a_{rl}^*}{x_r}, \cdots, \frac{a_{(u\cdot v-1),l}^*}{x_{u\cdot v-1}}, \frac{a_{(u\cdot v),l}^*}{x_{u\cdot v}}\right], (\neq A, r = \overline{1, u\cdot v}), l = 1, 2, \cdots, \text{ or simply,}$$

$$\widetilde{A}_l^* = \left[a_{1l}^*, \cdots, a_{rl}^*, \cdots, a_{(u\cdot v),l}^*\right] \quad (24)$$

$$\widetilde{B} = \left[\frac{b_1}{y_1}, \frac{b_2}{y_2}, \cdots, \frac{b_r}{y_r}, \cdots, \frac{b_{u\cdot v-1}}{y_{u\cdot v-1}}, \frac{b_{u\cdot v}}{y_{u\cdot v}}\right], (\neq B, r = \overline{1, u\cdot v}), \text{ or simply, } \widetilde{B} = \left[b_1, \cdots, b_r, \cdots, b_{u\cdot v}\right] \quad (25)$$

**Step 2**; Compute the extended distance measure $EDM(\widetilde{A}_l^*, \widetilde{A})$. Where index $l = 1, 2, \cdots$ is means the number of the given premise fuzzy set.

- Condition 1; When index $u > v$, then distance measure $EDM(\widetilde{A}_l^*, \widetilde{A})_\Theta$ ( $\Theta = \{u \text{ or } v \text{ or } u \cdot v\}$ ) is calculated as follows.

$$EDM(\widetilde{A}_l^*, \widetilde{A})_u = \left[\frac{1}{u}\sum_{i=1}^{u}\left[a_{il}^* - a_i\right]^2\right]^{1/2}, \text{ for } u > v, \ v = n \cdot u, \text{ FMP} \quad (26)$$

- Condition 2; When index $u < v$, then distance measure $DM(\widetilde{A}_l^*, \widetilde{A})$ is calculated as follows.

$$EDM(\widetilde{A}_l^*, \widetilde{A})_v = \left[\frac{1}{v}\sum_{j=1}^{v}\left[a_{jl}^* - a_j\right]^2\right]^{1/2}, \text{ for } u < v, \ u = n \cdot v, \text{ FMP} \quad (27)$$

- Condition 3; When index $u > v$, then distance measure $DM(\widetilde{A}_l^*, \widetilde{A})_\Theta$, ( $\Theta = \{u \text{ or } v \text{ or } u \cdot v\}$ ) is calculated as follows.

$$EDM(\widetilde{A}_l^*, \widetilde{A})_{u\cdot v} = \left[\frac{1}{u \cdot v}\sum_{r=1}^{u\cdot v}\left[a_{rl}^* - a_r\right]^2\right]^{1/2}, \text{ for } u \neq n \cdot v \text{ or } v \neq n \cdot u, \text{ FMP} \quad (28)$$

**Step 3**; Compute the sign vectors $\widetilde{P}_l$ by the difference $dif_{kl} = a_{kl}^* - a_k, (k = i \text{ or } j \text{ or } r, l = 1, 2, \cdots)$ of the given premise and the antecedent.

$$\widetilde{P}_l = \left[\widetilde{P}_{1l}, \widetilde{P}_{2l}, \cdots \widetilde{P}_{kl}, \cdots\right], \quad k = i \text{ or } j \text{ or } r, \quad l = 1, 2, \cdots \quad (29)$$

- *First P(+1,0,-1) form;* $\widetilde{P}_{kl} = sign(dif_{kl}) = \begin{cases} +1, & dif_{kl} > 0 \\ 0, & dif_{kl} = 0 \\ -1, & dif_{kl} < 0 \end{cases}$, for FMP-EDM (30)

- *Second P(+1,-1) form;* $\widetilde{P}_{kl} = sign(dif_{kl}) = \begin{cases} +1, & dif_{kl} \geq 0 \\ -1, & dif_{kl} < 0 \end{cases}$, for FMP-EDM (31)

**Step 4**; Compute the vectorial distance measure $\widetilde{C}_l$ since $EDM(\widetilde{A}_l^*, \widetilde{A})_\Theta$ with an index $\Theta = \{u \text{ or } v \text{ or } u \cdot v\}$ is a scalar.

$$\widetilde{C}_l = EDM(\widetilde{A}_l^*, \widetilde{A})_\Theta \times P_l, \ \Theta = \{u \text{ or } v \text{ or } u \cdot v\} \quad (32)$$

**Step 5**; Obtain the quasi-quasi-approximate reasoning results $\widetilde{B}_l^{**}, l = 1, 2, \cdots$ for FMP-EDM.

$$\tilde{B}_l^{**} = \begin{cases} \tilde{B}_l + \tilde{C}_l, & if \quad \text{Case 1, 2, and 3} \\ 1 - \tilde{B}_l + \tilde{C}_l, & if \quad \text{Case 4} \\ s.t.\ \tilde{B}_l + \tilde{C}_l, & if \quad \text{Case 5} \end{cases} \tag{33}$$

**Step 6**; Select the quasi-approximate reasoning results $\tilde{B}_l^*, l = 1, 2, \cdots$ from the quasi-quasi-approximate reasoning results $\tilde{B}_l^{**}$ for indexes $k = (i = \overline{1, u}$ or $j = \overline{1, v}$ or $r = \overline{1, u \cdot v})$, $l = 1, 2, \cdots$. We will call this FMP-EDM.

$$\tilde{B}_l^{**} \to \tilde{B}_l^*, \text{ i.e., } \left[\tilde{b}_{lk}^{**}\right]_{k\times 1} \to \left[\tilde{b}_{lj}^*\right]_{v\times 1} \tag{34}$$

Condition1; if the index $k = u$, i.e., $u > v$, $u = n \cdot v$, then $\tilde{b}_{li}^* = \tilde{b}_{l,(i\cdot n)}^{**}$

Condition2; if the index $k = v$, i.e., $u < v$, $v = n \cdot u$, then $\tilde{b}_{lj}^* = \tilde{b}_{lj}^{**}$

Condition3; if the index $k = u \cdot v$, i.e., $u \neq v$, then $\tilde{b}_{lr}^* = \tilde{b}_{l,(r\cdot u)}^{**}$

Matlab code for realization the formula (34) is as follows.

```
function [ outputV ] = re2_vector( inputV,u,v,fact )
    outputV = [;;;];
    for i = 1: u
        for j = 1: v
            outputV(i,j) = inputV(i, j * fact);
        end
    end
end
```

**Step 7**; Solve the individual approximate reasoning result $B_l^*$ from the quasi-approximate reasoning results $\tilde{B}_l^*$.

$$B_l^* = \frac{\tilde{B}_l^* - \eta_l}{\xi_l - \eta_l}, l = 1, 2, \cdots \tag{35}$$

Where, the maximum $\eta_l$ and minimum $\xi_l$ of $\tilde{B}_l^*$ for FMP is calculated as follows.

$$\eta_l = \max \tilde{B}_l^*, \xi_l = \min \tilde{B}_l^*, l = 1, 2, \cdots. \tag{36}$$

**Step 8**; For the SISO fuzzy system with discrete fuzzy set vectors of different dimensions, the final approximate reasoning result $B^*$ according to the given premises for FMP-EDM is obtained as follows.

$$B^* = \{B_l^*\}, \; l = 1, 2, \cdots \tag{37}$$

According to above mentioned method, let us consider following examples.

**Example 3.1.** About the proposed FMP-EDM, for 3×1 dimension antecedent fuzzy row vector *A*(*x*)=[1, 0.4, 0] and 4×1 dimension consequent fuzzy row vector *B*(*y*)=[0, 0.4, 0.7, 1]. When the given premise is $A^*(x)$ =*A*(*x*)=[1, 0.4, 0], let us consider FMP-EDM. The index is $u \neq n \cdot v, v \neq n \cdot u$, so $r = u \cdot v = 12$. The dimensions of the extended every vectors $\tilde{A}$, $\tilde{A}^*$, and $\tilde{B}$ are 12×1 dimension. The proposed approximate reasoning results are computed by two form, i.e., *P(+1, 0, -1) form* and *P(+1, -1) form.*

We compute the extended fuzzy row vectors $\tilde{A}$, $\tilde{A}^*$, and $\tilde{B}$.

$$\tilde{A} = \left[\frac{a_1}{x_1}, \frac{a_2}{x_2}, \cdots, \frac{a_r}{x_r}, \cdots, \frac{a_{11}}{x_{11}}, \frac{a_{12}}{x_{12}}\right] = [1, 1, 1, 1, 0.85, 0.7, 0.55, 0.4, 0.3, 0.2, 0.1, 0]$$

$$\tilde{A}^* = \left[\frac{a_1^*}{x_1}, \frac{a_2^*}{x_2}, \cdots, \frac{a_r^*}{x_r}, \cdots, \frac{a_{11}^*}{x_{11}}, \frac{a_{12}^*}{x_{12}}\right] = [1, 1, 1, 1, 0.85, 0.7, 0.55, 0.4, 0.3, 0.2, 0.1, 0]$$

$$\tilde{B} = \left[\frac{b_1}{y_1}, \frac{b_2}{y_2}, \cdots, \frac{b_r}{y_r}, \cdots, \frac{b_{11}}{y_{11}}, \frac{b_{12}}{y_{12}}\right] = [0, 0, 0, 0.1333, 0.2667, 0.4, 0.5, 0.6, 0.7, 0.8, 0.9, 1]$$

We compute the extended distance measure $EDM(\tilde{A}^*, \tilde{A})$.

$$EDM(\tilde{A}^*, \tilde{A})_{12} = \left[\frac{1}{12}\sum_{r=1}^{12}\left[a_r^* - a_r\right]^2\right]^{1/2} = 0$$

We compute the sign vectors $\tilde{P}$.

- *First P(+1, 0, -1) form;* $\tilde{P} = \left[\tilde{P}_1, \tilde{P}_2, \cdots \tilde{P}_{12}\right]$=[0, 0, 0, 0, 0, 0, 0, 0, 0, 0, 0, 0]
- *Second P(+1, -1) form;* $\tilde{P} = \left[\tilde{P}_1, \tilde{P}_2, \cdots \tilde{P}_{12}\right]$=[1, 1, 1, 1, 1, 1, 1, 1, 1, 1, 1, 1]

We compute the vectorial distance measure $\tilde{C}$.

- *First P(+1, 0, -1) form;* $\tilde{C} = EDM(\tilde{A}^*, \tilde{A})_{12} \times \tilde{P}$=[0, 0, 0, 0, 0, 0, 0, 0, 0, 0, 0, 0]

- *Second P(+1, -1) form;* $\widetilde{C} = EDM(\widetilde{A}^*, \widetilde{A})_{12} \times \widetilde{P}$=[0, 0, 0, 0, 0, 0, 0, 0, 0, 0, 0, 0]

We obtain the quasi-quasi-approximate reasoning results $\widetilde{B}^{**}$ for FMP-EDM. Then $\widetilde{A} = \widetilde{A}^*$ , therefore $\widetilde{B}^{**}$ is calculated as follows;

- *First P(+1, 0, -1) form;* $\widetilde{B}^{**} = \widetilde{B} + \widetilde{C} = \widetilde{B}$=[0, 0, 0, 0.1333, 0.2667, 0.4, 0.5, 0.6, 0.7, 0.8, 0.9, 1]
- *Second P(+1, -1) form;* $\widetilde{B}^{**} = \widetilde{B} + \widetilde{C} = \widetilde{B}$ =[0, 0, 0, 0.1333, 0.2667, 0.4, 0.5, 0.6, 0.7, 0.8, 0.9, 1]

We select the quasi-approximate reasoning results $\widetilde{B}^*$ from the quasi-quasi-approximate reasoning results $\widetilde{B}^{**}$.

$$\widetilde{B}^{**} \to \widetilde{B}^*, \text{ i.e., } \left[\widetilde{b}_k^{**}\right]_{k\times 1} \to \left[\widetilde{b}_j^*\right]_{v\times 1}$$

The index is $k = u \cdot v$ therefore $\widetilde{b}_{lr}^* = \widetilde{b}_{l,(r\cdot u)}^{**}$. The result is followed as;

- *First P(+1, 0, -1) form;* $\widetilde{B}^*$=[0, 0.4, 0.7, 1]
- *Second P(+1, -1) form;* $\widetilde{B}^*$=[0, 0.4, 0.7, 1]

We solve the individual approximate reasoning result $B^*$ from the quasi-approximate reasoning results $\widetilde{B}^*$.

- *P(+1, 0, -1) form;* $\eta = \max \widetilde{B}^* = 1, \xi = \min \widetilde{B}^* = 0$ , therefore $B^* = (\widetilde{B}^* - \eta)/(\xi - \eta) = \widetilde{B}^*$=[0, 0.4, 0.7, 1]
- *P(+1, -1) form;* $\eta = \max \widetilde{B}^* = 1, \xi = \min \widetilde{B}^* = 0$ , therefore $B^* = (\widetilde{B}^* - \eta)/(\xi - \eta) = \widetilde{B}^*$=[0, 0.4, 0.7, 1]

Therefore, the result is followed as;

- *First P(+1, 0, -1) form;* $B^*$=[0, 0.4, 0.7, 1]
- *Second P(+1, -1) form;* $B^*$=[0, 0.4, 0.7, 1]

**Example 3.2.** About the proposed FMP-EDM, For 4×1 dimension antecedent fuzzy row vector $A(x)$=[1, 0.8, 0.4, 0] and 6×1 dimension consequent fuzzy row vector $B(y)$=[0, 0.2, 0.4, 0.7, 0.9, 1]. When the given premise is $A^*(x)$=[1, 0.9, 0.3, 0], let us consider FMP-EDM. The index is $u \neq n\cdot v, v \neq n\cdot u$ , so $r = lcm(u,v) = 12$. The dimensions of the extended every vectors $\widetilde{A}$, $\widetilde{A}^*$, and $\widetilde{B}$ are 12×1 dimension. The proposed approximate reasoning results are computed by two form, i.e., *P(+1, 0, -1) form* and *P(+1, -1) form.*

We compute the extended fuzzy row vectors $\widetilde{A}$, $\widetilde{A}^*$, and $\widetilde{B}$.

$$\widetilde{A} = \left[\frac{a_1}{x_1}, \frac{a_2}{x_2}, \cdots, \frac{a_r}{x_r}, \cdots, \frac{a_{11}}{x_{11}}, \frac{a_{12}}{x_{12}}\right] = [1, 1, 1, 0.9333, 0.8667, 0.8, 0.6667, 0.5333, 0.4, 0.2667, 0.1333, 0]$$

$$\widetilde{A}^* = \left[\frac{a_1^*}{x_1}, \frac{a_2^*}{x_2}, \cdots, \frac{a_r^*}{x_r}, \cdots, \frac{a_{11}^*}{x_{11}}, \frac{a_{12}^*}{x_{12}}\right] = [1, 1, 1, 0.9667, 0.9333, 0.9, 0.7, 0.5, 0.3, 0.2, 0.1, 0]$$

$$\widetilde{B} = \left[\frac{b_1}{y_1}, \frac{b_2}{y_2}, \cdots, \frac{b_r}{y_r}, \cdots, \frac{b_{11}}{y_{11}}, \frac{b_{12}}{y_{12}}\right] = [0, 0, 0.1, 0.2, 0.3, 0.4, 0.55, 0.7, 0.8, 0.9, 0.95, 1]$$

We compute the extended distance measure $EDM(\widetilde{A}^*, \widetilde{A})$.

$$EDM(\widetilde{A}^*, \widetilde{A})_{12} = \left[\frac{1}{12}\sum_{r=1}^{12}\left[a_r^* - a_r\right]^2\right]^{1/2} = 0.0527$$

We compute the sign vectors $\widetilde{P}$.

- *First P(+1, 0, -1) form;* $\widetilde{P} = \left[\widetilde{P}_1, \widetilde{P}_2, \cdots \widetilde{P}_{12}\right]$= [0, 0, 0, 1, 1, 1, 1, -1, -1, -1, -1, 0]
- *Second P(+1, -1) form;* $\widetilde{P} = \left[\widetilde{P}_1, \widetilde{P}_2, \cdots \widetilde{P}_{12}\right]$=[1, 1, 1, 1, 1, 1, 1, -1, -1, -1, -1, 1]

We compute the vectorial distance measure $\widetilde{C}$.

- *P(+1, 0, -1) form;* $\widetilde{C} = EDM(\widetilde{A}^*, \widetilde{A})_{12} \times \widetilde{P} =$
  =[ 0, 0, 0, 0.0527, 0.0527, 0.0527, 0.0527, -0.0527, -0.0527, -0.0527, -0.0527, 0]
- *P(+1, -1) form;* $= \widetilde{C} = EDM(\widetilde{A}^*, \widetilde{A})_{12} \times \widetilde{P} =$
  =[0.0527, 0.0527, 0.0527, 0.0527, 0.0527, 0.0527, 0.05 27, -0.0527, -0.0527, -0.0527, -0.0527, 0.0527]

We obtain the quasi-quasi-approximate reasoning results $\widetilde{B}^{**}$ for FMP-EDM. Then $\widetilde{A}^* = s.t.\widetilde{A}$, therefore $\widetilde{B}^{**}$ is calculated as follows;

- *P(+1, 0, -1) form;* $\widetilde{B}^{**} = \widetilde{B} + \widetilde{C} =$
  =[0, 0, 0.1, 0.2527, 0.3527, 0.4527, 0.6027, 0.6473, 0.7473, 0.8473, 0.8973, 1]
- *P(+1, -1) form;* $\widetilde{B}^{**} = \widetilde{B} + \widetilde{C} =$
  =[0.0527, 0.0527, 0.1527, 0.2527, 0.3527, 0.4527, 0.6027 0.6473, 0.7473, 0.8473, 0.8973, 1.0527]

We select the quasi-approximate reasoning results $\widetilde{B}^*$ from the quasi-quasi-approximate reasoning results $\widetilde{B}^{**}$.

$$\widetilde{B}^{**} \to \widetilde{B}^*, \text{ i.e., } \left[\widetilde{b}_k^{**}\right]_{k\times 1} \to \left[\widetilde{b}_q^*\right]_{v\times 1}$$

The index is $k = u \cdot v$ therefore $\widetilde{b}^*_{lq} = \widetilde{b}^{**}_{l,(q \cdot u)}$. The computational result is followed as;

- *First P(+1, 0, -1) form;* $\widetilde{B}^*$=[0, 0.2527, 0.4527, 0.6473, 0.8473, 1]
- *Second P(+1, -1) form;* $\widetilde{B}^*$=[0.0527, 0.2527, 0.4527, 0.6473, 0.8473, 1.0527]

Step 7; We solve the individual approximate reasoning result $B^*$ from the quasi-approximate reasoning results $\widetilde{B}^*$.

- *First P(+1, 0, -1) form;* $\eta = \max \widetilde{B}^* = 1, \xi = \min \widetilde{B}^* = 0$, therefore

  $B^* = (\widetilde{B}^* - \eta)/(\xi - \eta) = \widetilde{B}^*$=[0, 0.2527, 0.4527, 0.6473, 0.8473, 1]
- *Second P(+1, -1) form;* $\eta = \max \widetilde{B}^* = 1.0527, \xi = \min \widetilde{B}^* = 0.0527$.

  Therefore, $B^* = (\widetilde{B}^* - \eta)/(\xi - \eta) = \widetilde{B}^*$=[0, 0.2, 0.4, 0.5946, 0.7946 , 1]

Therefore, the result is;

- *First P(+1, 0, -1) form;* $B^* = (\widetilde{B}^* - \eta)/(\xi - \eta) = \widetilde{B}^*$=[0, 0.2527, 0.4527, 0.6473, 0.8473, 1]
- *Second P(+1, -1) form;* $B^* = (\widetilde{B}^* - \eta)/(\xi - \eta) = \widetilde{B}^*$=[0, 0.2, 0.4, 0.5946, 0.7946 , 1]

**Theorem 3.1.** *For the SISO fuzzy system with different dimensions, if $A^*(x) = A(x) \subseteq F(X), x \in X$, and is applied FMP-EDM, then the reasoning result $B^*(y) \subseteq F(Y), y \in Y$ is the consequent $B(y) \subseteq F(Y), y \in Y$, thereby the reductive property is completely satisfied. Where $F(X), F(Y)$ are all the fuzzy subsets on the universe of discourse $X, Y$, respectively.*

**Proof.** Since the indexes $u, v, n$ are all real integer index, then from the extended fuzzy sets of the antecedent $A$, the given premise $A^*$ and consequent $B$, i.e., the extended fuzzy vectors $\widetilde{A}$, $\widetilde{A}^* = \{\widetilde{A}^*_l\}, \widetilde{A}^*_l = [\widetilde{a}^*_{1l}, \cdots, \widetilde{a}^*_{il}, \cdots, \widetilde{a}^*_{ul}], l = 1$, and $\widetilde{B}$, new fuzzy approximate reasoning results $B^*$ are calculated as following 3 conditions, respectively;

① When the index $u > v$, i.e., $u = n \cdot v$ then the difference between the extended every vectors $\widetilde{A}^*$ and $\widetilde{A}$ are calculated as follows; $\widetilde{A}^*_l - \widetilde{A} = \left[\frac{a^*_{1l} - a_1}{x_1}, \frac{a^*_{2l} - a_2}{x_2}, \cdots, \frac{a^*_{il} - a_i}{x_i}, \cdots, \frac{a^*_{u-1,l} - a_{u-1}}{x_{u-1}}, \frac{a^*_{u,l} - a_u}{x_u}\right]$, that is, $\widetilde{A}^*_l - \widetilde{A} = [a^*_{1l} - a_1, a^*_{2l} - a_2, \cdots, a^*_{il} - a_i, \cdots, a^*_{u-1,l} - a_{u-1}, a^*_{u,l} - a_u]$. Since $A^*(x) = A(x) \subseteq F(X), x \in X$, i.e., $a^*_{il} = a_i, i = \overline{1, u}$ then $\widetilde{A}^*_l - \widetilde{A} = 0$. Therefore, for every $u > v$, $u = n \cdot v$, the extended distance measure (EDM) $DM(\widetilde{A}^*_l, \widetilde{A})_u = \left[\frac{1}{u}\sum_{i=1}^{u}[a^*_{il} - a_i]^2\right]^{1/2} = 0$. Thereby the extended vectorial distance measure $\widetilde{C}_l = EDM(\widetilde{A}^*_l, \widetilde{A})_u \times P_l = 0$, the quasi-quasi-approximate reasoning results is obtained as $\widetilde{B}^{**}_l = \widetilde{B}_l + \widetilde{C}_l = \widetilde{B}_l$, and then the quasi- approximate reasoning results is computed as $\widetilde{B}^{**}_l \to \widetilde{B}^*_l$, i.e., $[\widetilde{b}^{**}_{lk}]_{k \times 1} \to [\widetilde{b}^*_{lj}]_{v \times 1}$. From the individual approximate reasoning result $B^*_l = \frac{\widetilde{B}^*_l - \eta_l}{\xi_l - \eta_l}, l = 1$, the final reasoning result is obtained as $B^* = \{B^*_l\} = B,\ l = 1$. Therefore ① of this theorem 3.1 is right.

② In the case of the index $u < v$, i.e., $v = n \cdot u$, the difference between the extended every vectors $\widetilde{A}^*$ and $\widetilde{A}$ are calculated as follows; $\widetilde{A}^*_l - \widetilde{A} = \left[\frac{a^*_{1l} - a_1}{x_1}, \frac{a^*_{2l} - a_2}{x_2}, \cdots, \frac{a^*_{ql} - a_q}{x_q}, \cdots, \frac{a^*_{n \cdot u-1,l} - a_{u-1}}{x_{n \cdot u-1}}, \frac{a^*_{n \cdot u,l} - a_{n \cdot u}}{x_{n \cdot u}}\right]$, for short, $\widetilde{A}^*_l - \widetilde{A} = [a^*_{1l} - a_1, a^*_{2l} - a_2, \cdots, a^*_{ql} - a_q, \cdots, a^*_{n \cdot u-1,l} - a_{n \cdot u-1}, a^*_{n \cdot u,l} - a_{n \cdot u}]$. From the condition of theorem 3.1, $A^*(x) = A(x) \subseteq F(X), x \in X$, i.e., $a^*_{ql} = a_q, q = \overline{1, n \cdot u}$, thus $\widetilde{A}^*_l - \widetilde{A} = 0$. Therefore, for every $u < v$, $v = n \cdot u$, the extended distance measure (EDM), $EDM(\widetilde{A}^*_l, \widetilde{A})_v = \left[\frac{1}{n \cdot u}\sum_{q=1}^{n \cdot u}[a^*_{ql} - a_q]^2\right]^{1/2} = 0$. Then the extended vectorial distance measure $\widetilde{C}_l = EDM(\widetilde{A}^*_l, \widetilde{A})_v \times P_l = 0$, then the quasi-quasi-approximate reasoning results is obtained as $\widetilde{B}^{**}_l = \widetilde{B}_l + \widetilde{C}_l = \widetilde{B}_l$, and then the quasi-approximate reasoning results is computed as $\widetilde{B}^{**}_l \to \widetilde{B}^*_l$, i.e., $[\widetilde{b}^{**}_{lk}]_{k \times 1} \to [\widetilde{b}^*_{lj}]_{v \times 1}$. From the individual approximate reasoning result $B^*_l = \frac{\widetilde{B}^*_l - \eta_l}{\xi_l - \eta_l}, l = 1$, the final reasoning result is obtained as $B^* = B = [b_1, \cdots, b_j, \cdots, b_v]$. Therefore ② of this theorem 1 is true.

③ The difference between the extended every vectors $\widetilde{A}^*$ and $\widetilde{A}$ according to the index $u \neq v$, are calculated as follows; $\widetilde{A}^*_l - \widetilde{A} = \left[\frac{a^*_{1l} - a_1}{x_1}, \frac{a^*_{2l} - a_2}{x_2}, \cdots, \frac{a^*_{rl} - a_r}{x_r}, \cdots, \frac{a^*_{u \cdot v-1,l} - a_{u \cdot v-1}}{x_{u \cdot v-1}}, \frac{a^*_{u \cdot v,l} - a_{u \cdot v}}{x_{u \cdot v}}\right]$, for short, $\widetilde{A}_l^* - \widetilde{A} = \left[a_{1l}^* - a_1, a_{2l}^* - a_2, \cdots, a_{rl}^* - a_r, \cdots, a_{n\cdot u-1,l}^* - a_{u\cdot v-1}, a_{u\cdot v,l}^* - a_{u\cdot v}\right]$. From the condition of theorem 1, since $a_{rl}^* = a_r, r = \overline{1, u\cdot v}$, following equation $\widetilde{A}_l^* - \widetilde{A} = 0$ is satisfied. Therefore, for every index $u \cdot v$, the extended distance measure (EDM), $EDM(\widetilde{A}_l^*, \widetilde{A})_{u\cdot v} = \left[\frac{1}{u\cdot v}\sum_{r=1}^{u\cdot v}\left[a_{rl}^* - a_r\right]^2\right]^{1/2} = 0$. Then the extended vectorial distance measure $\widetilde{C}_l = EDM(\widetilde{A}_l^*, \widetilde{A})_{u\cdot v} \times P_l = 0$, then the quasi-quasi-approximate reasoning results is obtained as $\widetilde{B}_l^{**} = \widetilde{B}_l + \widetilde{C}_l = \widetilde{B}_l$, and then the quasi-approximate reasoning results is computed as $\widetilde{B}_l^{**} \to \widetilde{B}_l^*$, i.e., $\left[\widetilde{b}_{l(u\cdot v)}^{**}\right]_{(u\cdot v)\times 1} \to \left[\widetilde{b}_{lj}^*\right]_{v\times 1}$. From the individual approximate reasoning result $B_l^* = \frac{\widetilde{B}_l^* - \eta_l}{\xi_l - \eta_l}, l = 1$, the final reasoning result is obtained as $B^* = B = \left[b_1, \cdots, b_j, \cdots, b_v\right]$. Hence ③ of this theorem 1 is right.

Therefore from ①, ②, ③, if the given premise discrete fuzzy vector is $A^*(x) = A(x) \subseteq F(X), x \in X$, and the proposed fuzzy approximate reasoning method, i.e., FMP-EDM is applied in the SISO fuzzy system, then the approximate reasoning result $B^*(y) \subseteq F(Y), y \in Y$ of FMP-EDM is obtained to equal to the consequent fuzzy vector $B(y) \subseteq F(Y), y \in Y$ of the fuzzy rule, thereby when $F(X), F(Y)$ are all the fuzzy subsets on the universe of discourse $X, Y$, respectively, the reductive property of the proposed FMP-EDM is completely satisfied. (Proof End)

**Experiment 3.1.** For the SISO fuzzy system with discrete fuzzy set vectors of different dimensions in Class 1, let us consider an approximate fuzzy reasoning based on 5×1 dimension antecedent fuzzy row vector $A(x)$=[1, 0.3, 0, 0, 0] and 7×1 dimension consequent fuzzy row vector $B(y)$=[0, 0, 0, 0, 0, 0.3, 1] for FMP-EDM. The given premises are $A_1^*(x) = A(x)$=[1, 0.3, 0, 0, 0], $A_2^*(x) = very\ A(x)$=[1, 0.09, 0, 0, 0], $A_3^*(x) = more\ or\ less\ A(x)$=[1, 0.55, 0, 0, 0], $A_4^*(x) = not\ A(x)$=[0, 0.7, 1, 1, 1], respectively. The proposed approximate reasoning results are computed by two form, i.e., *P(+1,0,-1) form* and *P(+1,-1) form*. The computational fuzzy approximate reasoning result by MATLB experiment is shown in Table 2.

**Table 2.** FMP-EDM Reasoning Results and Reductive Property in Class 1

| FMP-EDM Premise $A^*(x)$ | | FMP-EDM Reasoning Results and Reductive Property | | | |
|---|---|---|---|---|---|
| | | Reasoning Results $B^*(y)$ | | | *RPCF* |
| $A_1^*$ | [1, 0.3, 0, 0, 0] | $B_1^*$ | *P(+1,0,-1) form* | [0, 0, 0, 0, 0, 0.3, 1] | 100 (%) |
| | | | *P(+1,-1) form* | [0, 0, 0, 0, 0, 0.3, 1] | 100 (%) |
| $A_2^*$ | [1, 0.09, 0, 0, 0] | $B_2^*$ | *P(+1,0,-1) form* | [0.097, 0, 0, 0, 0.097, 0.37, 1] | 93.25 (%) |
| | | | *P(+1,-1) form* | [0.18, 0, 0, 0, 0.18, 0.42, 1] | 90.17 (%) |
| $A_3^*$ | [1, 0.55, 0, 0, 0] | $B_3^*$ | *P(+1,0,-1) form* | [0, 0.12, 0.12, 0.12, 0, 0.3, 1] | 91.21 (%) |
| | | | *P(+1,-1) form* | [0, 0, 0, 0, 0, 0.3, 1] | 96.46 (%) |
| $A_4^*$ | [0, 0.7, 1, 1, 1] | $B_4^*$ | *P(+1,0,-1) form* | [0, 0, 1, 1, 1, 0.83, 0.43] | 63.53 (%) |
| | | | *P(+1,-1) form* | [0, 0, 1, 1, 1, 0.83, 0.43] | 63.53 (%) |
| *RPCF* | | *FMP-EDM-P(+1,0,-1) form* | | | 87.00 (%) |
| | | *FMP-EDM-P(+1,-1) form* | | | 87.54 (%) |

**Experiment 3.2.** For the SISO fuzzy system with discrete fuzzy set vectors of different dimensions in Class 2, let us consider an approximate fuzzy reasoning based on 5×1 dimension antecedent fuzzy row vector $A(x)$=[1, 0.3, 0, 0, 0] and 7×1 dimension consequent fuzzy row vector $B(y)$=[0, 0, 0, 0, 0, 0.3, 1] for FMP-EDM. The given premises are $A_1^*(x) = A(x)$=[1, 0.3, 0, 0, 0], $A_2^*(x) = very\ A(x)$=[1, 0.09, 0, 0, 0], $A_3^*(x) = more\ or\ less\ A(x)$=[1, 0.55, 0, 0, 0], $A_4^*(x) = s.t.A(x)$=[1, 0.2, 0, 0, 0], respectively. The proposed fuzzy approximate reasoning results are computed by two form, i.e., *P(+1,0,-1) form* and *P(+1,-1) form*. The computational fuzzy approximate reasoning result by MATLB is shown in Table 3.

**Table 3.** FMP-EDM Reasoning Results and Reductive Property in Class 2

| FMP-EDM Premise $A^*(x)$ | | FMP-EDM Reasoning Results and Reductive Property | | | |
|---|---|---|---|---|---|
| | | Reasoning Results $B^*(y)$ | | | *RPCF* |
| $A_1^*$ | [1, 0.3, 0, 0, 0] | $B_1^*$ | *P(+1,0,-1) form* | [0, 0, 0, 0, 0, 0.3, 1] | 100 % |
| | | | *P(+1,-1) form* | [0, 0, 0, 0, 0, 0.3, 1] | 100 % |
| $A_2^*$ | [1, 0.09, 0, 0, 0] | $B_2^*$ | *P(+1,0,-1) form* | [0.097, 0, 0, 0, 0.097, 0.37, 1] | 93.25 % |
| | | | *P(+1,-1) form* | [0.18, 0, 0, 0, 0.18, 0.42, 1] | 90.17 % |
| $A_3^*$ | [1, 0.55, 0, 0, 0] | $B_3^*$ | *P(+1,0,-1) form* | [0, 0.12, 0.12, 0.12, 0, 0.3, 1] | 91.21 % |
| | | | *P(+1,-1) form* | [0, 0, 0, 0, 0, 0.3, 1] | 96.46 % |
| $A_4^*$ | [1, 0.2, 0, 0, 0] | $B_4^*$ | *P(+1,0,-1) form* | [0.035, 0, 0, 0, 0.035, 0.23, 1] | 98.58 % |
| | | | *P(+1,-1) form* | [0.068, 0, 0, 0, 0.068, 0.25, 1] | 97.26 % |
| *RPCF* | | *FMP-EDM-P(+1,0,-1) form* | | | 95.76 % |
| | | *FMP-EDM-P(+1,-1) form* | | | 95.97 % |

**Experiment 3.3.** For the SISO fuzzy system with discrete fuzzy set vectors of different dimensions in Class 1 and Class 2, let us consider the comprehensive reductive property of our proposed method EDM. (Table 4)

**Table 4.** Comprehension of our proposed FMP-EDM in Class 1 and Class 2

| | *FMP-EDM-P(+1,0,-1)form* | *FMP-EDM-P(+1,-1)form* |
|---|---|---|
| Class 1 | 87.00% | 87.54 % |
| Class 2 | 95.76 % | 95.97 % |
| RPCF-average | 91.38 % | 91.76% |

### 3.2. A Novel Fuzzy Approximate Reasoning Method For FMT

In this subsection, for the SISO fuzzy system with discrete fuzzy set vectors of different dimensions, we propose a novel FMT-DM method based on distance measure in the case that differs from dimensions between the antecedent and consequent, i.e., in the case of index $v \neq u$ when element number of the antecedent is *v* and element number of the consequent is *u*.

Let us promise several concepts of FMP for an approximate reasoning in SISO fuzzy system with discrete fuzzy set vectors of different dimensions.

- Let $Y$ be universe of discourse and $\overline{B}$ $(\overline{B} \subseteq Y)$ be membership function of the antecedent fuzzy set of $Y$, then $\overline{B} = [1-b_j]_{v\times 1},\ j=\overline{1,v}$ is called a $v\times 1$ dimension antecedent row vector.
- Let $X$ be universe of discourse and $\overline{A}$ $(\overline{A} \subseteq X)$ be membership function of the consequent fuzzy set of $X$, then $\overline{A} = [1-a_i]_{u\times 1},\ i=\overline{1,u}$ is called a $u\times 1$ dimension consequent row vector.
- Let $Y$ be universe of discourse and $B^* = \{B_l^*\}$ $(B_l^* \subseteq Y, l=1,2,\cdots)$ be membership function of the given premise fuzzy set of $Y$, then $B_l^* = [b_{jl}^*]_{v\times 1},\ j=\overline{1,v}, l=1,2,\cdots$ is called a $v\times 1$ dimension premise row vector.
- Let $X$ be universe of discourse and $A^* = \{A_l^*\}$ $(A_l^* \subseteq X, l=1,2,\cdots)$ be membership function of the conclusion fuzzy set of $X$, then $A_l^* = [a_{il}^*]_{u\times 1}, i=\overline{1,u}, l=1,2,\cdots$ is called a $u\times 1$ dimension conclusion row vector.
- Among fuzzy sets $\overline{B}, B^*, \overline{A}\ and\ A^*$, its elements, i.e., values of membership functions $1-b_j \in \overline{B}, b_{jl}^* \in B_l^*, j=\overline{1,v}, 1-a_i \in \overline{A}, and\ a_{il}^* \in A_l^*, i=\overline{1,u}$ are all the values of closed interval [0, 1] that characterizes the antecedent, the given premise, the consequent and new conclusion obtained by approximate reasoning, respectively.
- Every the antecedent, the given premise, the consequent and new conclusion fuzzy row vectors can be expressed as follows; $\overline{B} = [1-b_1,\cdots,1-b_j,\cdots,1-b_v]$ , $B^* = \{B_l^*\}, B_l^* = [b_{1l}^*,\cdots,b_{jl}^*,\cdots,b_{vl}^*], l=1,2,\cdots,$ , $\overline{A} = [1-a_1,\cdots,1-a_i,\cdots,1-a_u]$, and $A^* = \{A_l^*\}, A_l^* = [a_{1l}^*,\cdots,a_{il}^*,\cdots,a_{ul}^*], l=1,2,\cdots,$ respectively.

A novel approximate reasoning method of FMT for SISO fuzzy system with discrete fuzzy set vectors of different dimensions is described as following steps, which is so called FMT-EDM.

**Step 1**; Compute the extended fuzzy row vectors.

Let $u, v, n$ be all real integer index, then the extended fuzzy sets of the antecedent $\overline{B}$, the given premise $B^*$ and consequent $\overline{A}$, i.e., the extended fuzzy vectors $\tilde{\overline{B}}$, $\tilde{B}^* = \{\tilde{B}_l^*\}, \tilde{B}_l^* = [\tilde{b}_{1l}^*,\cdots,\tilde{b}_{il}^*,\cdots,\tilde{b}_{ul}^*], l=1,2,\cdots, and\ \tilde{\overline{A}}$ are calculated as following 3 conditions, respectively;

- Condition 1; If index $u < v$, i.e., $u = n\cdot v$ then the extended every vectors $\tilde{\overline{A}}$, $\tilde{\overline{B}}$, and $\tilde{B}^*$ are calculated as follows, respectively.

$$\tilde{\overline{A}} = \overline{A} = \left[\frac{1-a_1}{x_1}, \frac{1-a_2}{x_2}, \cdots, \frac{1-a_p}{x_p}, \cdots, \frac{1-a_{u-1}}{x_{u-1}}, \frac{1-a_u}{x_u}\right],$$

or simply, $\tilde{\overline{A}} = \overline{A} = [1-a_1, 1-a_2, \cdots, 1-a_{u-1}, 1-a_u]$ (38)

$$\tilde{\overline{B}} = \left[\frac{1-b_1}{y_1}, \frac{1-b_2}{y_2}, \cdots, \frac{1-b_i}{y_i}, \cdots, \frac{1-b_{n\cdot v-1}}{y_{n\cdot v-1}}, \frac{1-b_{n\cdot v}}{y_{n\cdot v}}\right], (\neq B, i=\overline{1,n\cdot v}),$$ or simply,

$$\tilde{\overline{B}} = [1-b_1, 1-b_2, \cdots, 1-b_{n\cdot v-1}, 1-b_{n\cdot v}] \quad (39)$$

$$\widetilde{\widetilde{B}}_l^* = \left[ \frac{b_{1l}^*}{y_1}, \frac{b_{2l}^*}{y_2}, \cdots, \frac{b_{il}^*}{y_i}, \cdots, \frac{b_{(n\cdot v-1),l}^*}{y_{n\cdot v-1}}, \frac{b_{(n\cdot v),l}^*}{y_{n\cdot v}} \right], (\neq B^*, i = \overline{1, n\cdot v}), l = 1,2,\cdots,$$

or simply, $\widetilde{B}_l^* = \left[ b_{1l}^*, b_{2l}^*, \cdots, b_{pl}^*, \cdots, b_{(n\cdot v-1),l}^*, b_{(n\cdot v),l}^* \right]$

- Condition 2; If index $u < v$, i.e., $v = n\cdot u$ then the extended every vectors $\widetilde{\widetilde{A}}$, $\widetilde{\widetilde{B}}$, and $\widetilde{B}^*$ are calculated as follows, respectively.

$$\widetilde{\widetilde{A}} = \left[ \frac{1-a_1}{x_1}, \frac{1-a_2}{x_2} \cdots, \frac{1-a_q}{x_q}, \cdots, \frac{1-a_{n\cdot u-1}}{x_{n\cdot u-1}}, \frac{1-a_{n\cdot u}}{x_{n\cdot u}} \right], (\neq \overline{A}, q = \overline{1, n\cdot u}),$$

or simply, $\widetilde{\widetilde{A}} = \left[ 1-a_1, \cdots, 1-a_q, \cdots, 1-a_{n\cdot u} \right]$ (40)

$$\widetilde{\widetilde{B}} = \overline{B} = \left[ \frac{1-b_1}{y_1}, \frac{1-b_2}{y_2}, \cdots, \frac{1-b_j}{y_j}, \cdots, \frac{1-b_{v-1}}{y_{v-1}}, \frac{1-b_v}{y_v} \right],$$

or simply, $\widetilde{\widetilde{B}} = \overline{B} = \left[ 1-b_1, \cdots, 1-b_j, \cdots, 1-b_v \right]$ (41)

$$\widetilde{B}_l^* = B_l^* = \left[ \frac{b_{1l}^*}{y_1}, \frac{b_{2l}^*}{y_2}, \cdots, \frac{b_{jl}^*}{y_j}, \cdots, \frac{b_{v-1,l}^*}{y_{v-1}}, \frac{b_{v,l}^*}{y_v} \right], \text{ or simply, } \widetilde{B}_l^* = B_l^* = \left[ b_{1l}^*, \cdots, b_{jl}^*, \cdots, b_{vl}^* \right] \quad (42)$$

- Condition 3; If index $u \neq v$, i.e., $u \neq n\cdot v$ or $v \neq n\cdot u$ then the extended every vectors $\widetilde{\widetilde{A}}$, $\widetilde{\widetilde{B}}$, and $\widetilde{B}^*$ are calculated as follows, respectively.

$$\widetilde{\widetilde{A}} = \left[ \frac{1-a_1}{x_1}, \frac{1-a_2}{x_2}, \cdots, \frac{1-a_r}{x_r}, \cdots, \frac{1-a_{u\cdot v-1}}{x_{u\cdot v-1}}, \frac{1-a_{u\cdot v}}{x_{u\cdot v}} \right], (\neq \overline{A}, r = \overline{1, u\cdot v}),$$

or simply, $\widetilde{\widetilde{A}} = \left[ 1-a_1, \cdots, 1-a_r, \cdots, 1-a_{u\cdot v} \right]$ (43)

$$\widetilde{\widetilde{B}} = \left[ \frac{1-b_1}{y_1}, \frac{1-b_2}{y_2}, \cdots, \frac{1-b_r}{y_r}, \cdots, \frac{1-b_{u\cdot v-1}}{y_{u\cdot v-1}}, \frac{1-b_{u\cdot v}}{y_{u\cdot v}} \right], (\neq \overline{B}, r = \overline{1, u\cdot v}),$$

or simply, $\overline{B} = \left[ 1-b_1, \cdots, 1-b_r, \cdots, 1-b_{u\cdot v} \right]$ (44)

$$\widetilde{B}_l^* = \left[ \frac{b_{1l}^*}{y_1}, \frac{b_{2l}^*}{y_2}, \cdots, \frac{b_{rl}^*}{y_r}, \cdots, \frac{b_{(u\cdot v-1),l}^*}{y_{u\cdot v-1}}, \frac{b_{(u\cdot v),l}^*}{y_{u\cdot v}} \right], (\neq B^*, r = \overline{1, u\cdot v}), l = 1,2,\cdots,$$

or simply, $\widetilde{B}_l^* = \left[ b_{1l}^*, \cdots, b_{rl}^*, \cdots, b_{u\cdot v,l}^* \right]$ (45)

**Step 2**; Compute the extended distance measure $EDM(\widetilde{B}_l^*, \widetilde{\widetilde{B}})$. Where index $l = 1, 2, \cdots$ means the number of the given premise fuzzy set.

- Condition 1; When index $u > v$, then distance measure $EDM(\widetilde{B}_l^*, \widetilde{\widetilde{B}})_\Theta (\Theta = \{u\, or\, v\, or\, u\cdot v\})$ is calculated as follows.

$$EDM(\widetilde{B}_l^*, \widetilde{\widetilde{B}})_u = \left[ \frac{1}{u} \sum_{i=1}^{u} \left[ b_{il}^* - (1-b_i) \right]^2 \right]^{1/2}, \text{ for } u > v, \ v = n\cdot u, \text{ FMT} \quad (46)$$

- Condition 2; When index $u < v$, then distance measure $EDM(\widetilde{B}_l^*, \widetilde{\widetilde{B}})_\Theta (\Theta = \{u\, or\, v\, or\, u\cdot v\})$ is calculated as follows.

$$EDM(\widetilde{B}_l^*, \widetilde{\widetilde{B}})_v = \left[ \frac{1}{v} \sum_{j=1}^{v} \left[ b_{jl}^* - (1-b_j) \right]^2 \right]^{1/2}, \text{ for } u < v, \ u = n\cdot v, \text{ FMT} \quad (47)$$

- Condition 3; When index $u \neq v$, then distance measure $EDM(\widetilde{B}_l^*, \widetilde{\widetilde{B}})_\Theta (\Theta = \{u\, or\, v\, or\, u\cdot v\})$ is calculated as follows.

$$EDM(\widetilde{B}_l^*, \widetilde{\widetilde{B}})_{u\cdot v} = \left[ \frac{1}{u\cdot v} \sum_{r=1}^{u\cdot v} \left[ b_{rl}^* - (1-b_r) \right]^2 \right]^{1/2}, \text{ for } u \neq n\cdot v \text{ or } v \neq n\cdot u, \text{ FMT} \quad (48)$$

**Step 3**; Compute the sign vectors $\widetilde{P}_l$ by the difference $dif_{kl} = b_{kl}^* - (1-b_k), (k = i \text{ or } j \text{ or } r, l = 1,2,\cdots)$ of the given premise and the antecedent.

$$\widetilde{P}_l = \left[ \widetilde{P}_{1l}, \widetilde{P}_{2l}, \cdots \widetilde{P}_{kl}, \cdots \right], \quad k = i \text{ or } j \text{ or } r, \quad l = 1, 2, \cdots \quad (49)$$

- *First P(+1,0,-1) form;* $\widetilde{P}_{kl} = sign(dif_{kl}) = \begin{cases} +1, & dif_{kl} > 0 \\ 0, & dif_{kl} = 0 \\ -1, & dif_{kl} < 0 \end{cases}$, for FMT-EDM (50)

- *Second P(+1,-1) form;* $\widetilde{P}_{kl} = sign(dif_{kl}) = \begin{cases} +1, & dif_{kl} \geq 0 \\ -1, & dif_{kl} < 0 \end{cases}$, for FMT-EDM (51)

**Step 4**; Compute the vectorial distance measure $\widetilde{C}_l$ since $EDM(\widetilde{B}_l^*, \widetilde{\widetilde{B}})_\Theta$ with an index $\Theta = \{u \; or \; v \; or \; u \cdot v\}$ is a scalar.

$$\widetilde{C}_l = EDM(\widetilde{B}_l^*, \widetilde{\widetilde{B}})_\Theta \times P_l, \; \Theta = \{u \; or \; v \; or \; u \cdot v\} \tag{52}$$

**Step 5**; Obtain the quasi-quasi-approximate reasoning results $\widetilde{A}_l^{**}, l = 1,2,\cdots$ for FMT-EDM.

$$\widetilde{A}_l^{**} = \begin{cases} \widetilde{\widetilde{A}}_l + \widetilde{C}_l, & if \;\; \text{Case 6, 7, and 8} \\ \widetilde{A}_l + \widetilde{C}_l, & if \;\; \text{Case 9} \\ s.t. \; \widetilde{A}_l + \widetilde{C}_l, & if \;\; \text{Case 10} \end{cases} \tag{53}$$

**Step 6**; Select the quasi-approximate reasoning results $\widetilde{A}_l^*, l = 1,2,\cdots$ from the quasi-quasi-approximate reasoning results $\widetilde{A}_l^{**}$ for indexes $k = (i = \overline{1,u}$ or $j = \overline{1,v}$ or $r = \overline{1,u \cdot v})$, $l = 1, 2, \cdots$. We will call this FMT-EDM.

$$\widetilde{A}_l^{**} \to \widetilde{A}_l^*, \text{ i.e., } [\widetilde{a}_{lk}^{**}]_{k\times 1} \to [\widetilde{a}_{lq}^{*}]_{u\times 1} \tag{54}$$

Condition1; if the index $k = u$, i.e., $u > v$, $u = n \cdot v$, then $\widetilde{a}_{li}^* = \widetilde{a}_{li}^{**}$

Condition2; if the index $k = v$, i.e., $u < v$, $v = n \cdot u$, then $\widetilde{a}_{lj}^* = \widetilde{a}_{l,(j \cdot n)}^{**}$

Condition3; if the index $k = u \cdot v$, i.e., $u \neq v$, then $\widetilde{a}_{lr}^* = \widetilde{a}_{l,(r \cdot v)}^{**}$

**Step 7**; Solve the individual approximate reasoning result $A_l^*$ from the quasi-approximate reasoning results $\widetilde{A}_l^*$.

$$A_l^* = \frac{\widetilde{A}_l^* - \eta_l}{\xi_l - \eta_l}, l = 1, 2, \cdots \tag{55}$$

Where, the maximum $\eta_l$ and minimum $\xi_l$ of $\widetilde{A}_l^*$ is calculated as follows.

$$\eta_l = \max \widetilde{A}_l^*, \xi_l = \min \widetilde{A}_l^*. \tag{56}$$

**Step 8**; For the SISO fuzzy system with discrete fuzzy set vectors of different dimensions, the final approximate reasoning result $A^*$ according to the given premises for FMT-EDM is obtained as follows.

$$A^* = \{A_l^*\}, \; l = 1, 2, \cdots \tag{57}$$

**Example 3.3.** About the proposed FMT-EDM, for 4×1 dimension antecedent fuzzy row vector 1-*B*(*y*)=[1, 0.6, 0.3, 0] and3×1 dimension consequent fuzzy row vector 1-*A*(*x*)=[0, 0.6, 1]. When the given premise is $B^*(y)$=[1, 0.6, 0.3, 0], let us consider FMT-EDM. The index is $u \neq n \cdot v, v \neq n \cdot u$, so $r = u \cdot v = 12$. The dimensions of the extended every vectors $\widetilde{\widetilde{A}}$, $\widetilde{\widetilde{B}}$, and $\widetilde{B}^*$ are 12×1 dimension. Therefore the proposed approximate reasoning results for FMT-EDM are computed by two form, i.e., *P(+1, 0, -1) form* and *P(+1, -1) form*.

We compute the extended fuzzy row vectors $\widetilde{\widetilde{A}}$, $\widetilde{\widetilde{B}}$, and $\widetilde{B}^*$.

$$\widetilde{\widetilde{A}} = \left[\frac{1-a_1}{x_1}, \frac{1-a_2}{x_2}, \dots, \frac{1-a_r}{x_r}, \dots, \frac{1-a_{11}}{x_{11}}, \frac{1-a_{12}}{x_{12}}\right] = [0, 0, 0, 0, 0.15, 0.3, 0.45, 0.6, 0.7, 0.8, 0.9, 1]$$

$$\widetilde{\widetilde{B}} = \left[\frac{1-b_1}{y_1}, \frac{1-b_2}{y_2}, \dots, \frac{1-b_r}{y_r}, \dots, \frac{1-b_{11}}{y_{11}}, \frac{1-b_{12}}{y_{12}}\right] = [1, 1, 1, 0.8667, 0.7333, 0.6, 0.5, 0.4, 0.3, 0.2, 0.1, 0]$$

$$\widetilde{B}_l^* = \left[\frac{b_1^*}{y_1}, \frac{b_2^*}{y_2}, \dots, \frac{b_r^*}{y_r}, \dots, \frac{b_{11}^*}{y_{11}}, \frac{b_{12}^*}{y_{12}}\right] = [1, 1, 1, 0.8667, 0.7333, 0.6, 0.5, 0.4, 0.3, 0.2, 0.1, 0]$$

We compute the extended distance measure $EDM(\widetilde{B}^*, \widetilde{\widetilde{B}})$.

$$EDM(\widetilde{B}^*, \widetilde{\widetilde{B}})_{12} = \left[\frac{1}{12}\sum_{r=1}^{12}\left[b_r^* - (1-b_r)\right]^2\right]^{1/2} = 0$$

We compute the sign vectors $\widetilde{P}$.

- *First P(+1, 0, -1) form;* $\widetilde{P} = [\widetilde{P}_1, \widetilde{P}_2, \cdots \widetilde{P}_{12}]$=[0, 0, 0, 0, 0, 0, 0, 0, 0, 0, 0, 0]
- *Second P(+1, -1) form;* $\widetilde{P} = [\widetilde{P}_1, \widetilde{P}_2, \cdots \widetilde{P}_{12}]$=[1, 1, 1, 1, 1, 1, 1, 1, 1, 1, 1, 1]

We compute the vectorial distance measure $\widetilde{C}$.

- *First P(+1, 0, -1) form;* $\widetilde{C} = EDM(\widetilde{B}^*, \widetilde{\widetilde{B}})_{12} \times P$=[0, 0, 0, 0, 0, 0, 0, 0, 0, 0, 0, 0]

- *Second P(+1, -1) form;* $\widetilde{C} = EDM(\widetilde{B}^*, \widetilde{\widetilde{B}})_{12} \times P$=[0, 0, 0, 0, 0, 0, 0, 0, 0, 0, 0, 0]

We obtain the quasi-quasi-approximate reasoning results $\widetilde{A}^{**}$ for FMT-EDM.

Then $\widetilde{\widetilde{B}} = \widetilde{B}^*$ , therefore $\widetilde{A}^{**}$ is calculated as follows;

- *First P(+1, 0, -1) form;* $\widetilde{A}^{**} = \widetilde{\widetilde{A}} + \widetilde{C} = \widetilde{\widetilde{A}}$=[0, 0, 0, 0, 0.15, 0.3, 0.45, 0.6, 0.7, 0.8, 0.9, 1]
- *Second P(+1, -1) form;* $\widetilde{A}^{**} = \widetilde{\widetilde{A}} + \widetilde{C} = \widetilde{\widetilde{A}}$ = [0, 0, 0, 0, 0.15, 0.3, 0.45, 0.6, 0.7, 0.8, 0.9, 1]

We select the quasi-approximate reasoning results $\widetilde{A}^*$ from the quasi-quasi-approximate reasoning results $\widetilde{A}^{**}$.

$$\widetilde{A}^{**} \to \widetilde{A}^*, \text{ i.e., } \left[\widetilde{a}_k^{**}\right]_{k\times 1} \to \left[\widetilde{a}_p^*\right]_{u\times 1}$$

The index is $k = u \cdot v$ therefore $\widetilde{a}_p^* = \widetilde{a}_{p\cdot v}^{**}$. The result is followed as;

- *First P(+1, 0, -1) form;* $\widetilde{A}^*$=[0, 0.6, 1]
- *Second P(+1, -1) form;* $\widetilde{A}^*$=[0, 0.6, 1]

Step 7; We solve the individual approximate reasoning result $A^*$ from the quasi-approximate reasoning results $\widetilde{A}^*$.

- *P(+1, 0, -1) form;* $\eta = \max \widetilde{A}^* = 1, \xi = \min \widetilde{A}^* = 0$, therefore $A^* = (\widetilde{A}^* - \eta)/(\xi - \eta) = \widetilde{A}^*$=[0, 0.6, 1]
- *P(+1, -1) form;* $\eta = \max \widetilde{A}^* = 1, \xi = \min \widetilde{A}^* = 0$, therefore $A^* = (\widetilde{A}^* - \eta)/(\xi - \eta) = \widetilde{A}^*$=[0, 0.6, 1]

Therefore, the result is followed as;

- *First P(+1, 0, -1) form;* $A^*$=[0, 0.6, 1]
- *Second P(+1, -1) form;* $A^*$=[0, 0.6, 1]

**Example 3.4.** About the proposed FMT-EDM, for 6×1 dimension antecedent fuzzy row vector 1-*B*(*y*)=[1, 0.8, 0.6, 0.3, 0.1, 0] and 4×1 dimension consequent fuzzy row vector 1-*A*(*x*)=[0, 0.2, 0.6, 1]. When the given premise is $B^*(y)$=[1, 0.9, 0.8, 0.3, 0.1, 0], let us consider FMP-EDM. The proposed approximate reasoning results are computed by two form, i.e., *P(+1, 0, -1) form* and *P(+1, -1) form*. The index is $u \neq n\cdot v, v \neq n\cdot u$, so $r = u\cdot v = 12$. The dimensions of the extended every vectors $\widetilde{\widetilde{A}}$, $\widetilde{\widetilde{B}}$, and $\widetilde{B}^*$ are 12×1 dimension. Therefore the proposed approximate reasoning results for FMT-EDM are computed by two form, i.e., *P(+1, 0, -1) form* and *P(+1, -1) form.*

We compute the extended fuzzy row vectors $\widetilde{\widetilde{A}}$, $\widetilde{\widetilde{B}}$, and $\widetilde{B}^*$.

$$\widetilde{\widetilde{A}} = \left[\frac{1-a_1}{x_1}, \frac{1-a_2}{x_2}, \dots, \frac{1-a_r}{x_r}, \dots, \frac{1-a_{11}}{x_{11}}, \frac{1-a_{12}}{x_{12}}\right] = [0, 0, 0, 0.0667, 0.1333, 0.2, 0.3333, 0.4667, 0.6, 0.7333, 0.8667, 1]$$

$$\widetilde{\widetilde{B}} = \left[\frac{1-b_1}{y_1}, \frac{1-b_2}{y_2}, \dots, \frac{1-b_r}{y_r}, \dots, \frac{1-b_{11}}{y_{11}}, \frac{1-b_{12}}{y_{12}}\right] = [1, 1, 0.9, 0.8, 0.7, 0.6, 0.45, 0.3, 0.2, 0.1, 0.05, 0]$$

$$\widetilde{B}_l^* = \left[\frac{b_1^*}{y_1}, \frac{b_2^*}{y_2}, \dots, \frac{b_r^*}{y_r}, \dots, \frac{b_{11}^*}{y_{11}}, \frac{b_{12}^*}{y_{12}}\right] = [1, 1, 0.95, 0.9, 0.85, 0.8, 0.55, 0.3, 0.2, 0.1, 0.05, 0]$$

We compute the extended distance measure $EDM(\widetilde{B}^*, \widetilde{\widetilde{B}})$.

$$EDM(\widetilde{B}^*, \widetilde{\widetilde{B}})_{12} = \left[\frac{1}{12}\sum_{r=1}^{12}\left[b_r^* - (1-b_r)\right]^2\right]^{1/2} = 0.0842$$

We compute the sign vectors $\widetilde{P}$.

- *First P(+1, 0, -1) form;* $\widetilde{P} = \left[\widetilde{P}_1, \widetilde{P}_2, \cdots \widetilde{P}_{12}\right]$=[0, 0, 1, 1, 1, 1, 1, 0, 0, 0, 0, 0]
- *Second P(+1, -1) form;* $\widetilde{P} = \left[\widetilde{P}_1, \widetilde{P}_2, \cdots \widetilde{P}_{12}\right]$=[1, 1, 1, 1, 1, 1, 1, 1, 1, 1, 1, 1]

We compute the vectorial distance measure $\widetilde{C}$.

- *First P(+1, 0, -1) form;* $\widetilde{C} = EDM(\widetilde{B}^*, \widetilde{\widetilde{B}})_{12} \times P$=[0, 0, 0.0842, 0.0842, 0.0842, 0.0842, 0.0842, 0, 0, 0, 0, 0]
- *Second P(+1, -1) form;* $\widetilde{C} = EDM(\widetilde{B}^*, \widetilde{\widetilde{B}})_{12} \times P$=[0.0842, 0.0842, 0.0842, 0.0842, 0.0842, 0.0842, 0.0842, 0.0842, 0.0842, 0.0842, 0.0842, 0.0842]

We obtain the quasi-quasi-approximate reasoning results $\widetilde{A}^{**}$ for FMT-EDM. Then $\widetilde{B}^* = s.t.\widetilde{\widetilde{B}}$ , therefore $\widetilde{A}^{**}$ is calculated as follows;

- *First P(+1, 0, -1) form;* $\widetilde{A}^{**} = \widetilde{\widetilde{A}} + \widetilde{C} = \widetilde{\widetilde{A}} =$

  =[0, 0, 0.0842, 0.1508, 0.2175, 0.2842, 0.4175, 0.4667, 0.6, 0.7333, 0.8667, 1]
- *Second P(+1, -1) form;* $\widetilde{A}^{**} = \widetilde{\widetilde{A}} + \widetilde{C} = \widetilde{\widetilde{A}}$ = [0.0842, 0.0842, 0.0842, 0.1508, 0.2175, 0.2842, 0.4175, 0.5508,

0.6842, 0.8175, 0.9509, 1.0842]

We select the quasi-approximate reasoning results $\widetilde{A}^*$ from the quasi-quasi-approximate reasoning results $\widetilde{A}^{**}$.

$$\widetilde{A}^{**} \to \widetilde{A}^*, \text{ i.e., } [\widetilde{a}_k^{**}]_{k\times 1} \to [\widetilde{a}_p^*]_{u\times 1}$$

The index is $k = u \cdot v$ therefore $\widetilde{a}_p^* = \widetilde{a}_{p\cdot v}^{**}$. The result is followed as;

- *First P(+1, 0, -1) form;* $\widetilde{A}^*$=[0.0842, 0.2842, 0.6, 1]
- *Second P(+1, -1) form;* $\widetilde{A}^*$=[0.0842, 0.2842, 0.6842, 1.0842]

Step 7; We solve the individual approximate reasoning result $A^*$ from the quasi-approximate reasoning results $\widetilde{A}^*$.

- *P(+1, 0, -1) form;* $\eta = \max \widetilde{A}^* = 1, \xi = \min \widetilde{A}^* = 0.0842$, hence $A^* = (\widetilde{A}^* - \eta)/(\xi - \eta)$=[0, 0.2184, 0.5632, 1]
- *P(+1, -1) form;* $\eta = \max \widetilde{A}^* = 1.0842, \xi = \min \widetilde{A}^* = 0.0842$, therefore $A^* = (\widetilde{A}^* - \eta)/(\xi - \eta)$=[0, 0.2, 0.6, 1]

Therefore, the result is followed as;

- *First P(+1, 0, -1) form;* $A^*$=[0, 0.2184, 0.5632, 1]
- *Second P(+1, -1) form;* $A^*$=[0, 0.2, 0.6, 1]

**Theorem 3.2.** *For the SISO fuzzy system with different dimensions, if $B^*(y) = \overline{B}(y) \subseteq F(Y), y \in Y$, and is applied FMT-EDM, then the reasoning result $A^*(x) \subseteq F(X), x \in X$ is the consequent $\overline{A}(x) \subseteq F(X), x \in X$, thereby the reductive property is completely satisfied. Where $F(X), F(Y)$ are all the fuzzy subsets on the universe of discourse $X, Y$, respectively.*

**Proof.** Since the indexes $u, v, n$ are all real integer index, then from the extended fuzzy sets of the antecedent $\overline{B}$, the given premise $B^*$ and consequent $\overline{A}$, i.e., the extended fuzzy vectors $\widetilde{\overline{B}}$, $\widetilde{B}^* = \{\widetilde{B}_l^*\}, \widetilde{B}_l^* = [\widetilde{b}_{1l}^*, \cdots, \widetilde{b}_{il}^*, \cdots, \widetilde{b}_{ul}^*], l = 1$, and $\widetilde{\overline{A}}$, new fuzzy approximate reasoning results $A^*$ are calculated as following 3 conditions, respectively;

① In the case of the index $u > v$, i.e., $u = n \cdot v$, the difference between the extended every vectors $\widetilde{B}^*$ and $\widetilde{\overline{B}}$ are calculated as follows;

$$\widetilde{B}_l^* - \widetilde{\overline{B}} = \left[\frac{b_{1l}^* - (1-b_1)}{y_1}, \frac{b_{2l}^* - (1-b_2)}{y_2}, \cdots, \frac{b_{il}^* - (1-b_i)}{y_i}, \cdots, \frac{b_{n\cdot v-1,l}^* - (1-b_{n\cdot v-1})}{y_{n\cdot v-1}}, \frac{b_{n\cdot v,l}^* - (1-b_{n\cdot v})}{y_{n\cdot v}}\right]$$, for short,

$\widetilde{B}_l^* - \widetilde{\overline{B}} = [b_{1l}^* - (1-b_1), b_{2l}^* - (1-b_2), \cdots, b_{il}^* - (1-b_i), \cdots, b_{n\cdot v-1,l}^* - (1-b_{n\cdot v-1}), b_{n\cdot v,l}^* - (1-b_{n\cdot v})]$. From the condition of theorem 3.2, $B^*(y) = \overline{B}(y) \subseteq F(Y), y \in Y$, i.e., $b_{il}^* = 1 - b_i, i = \overline{1, n\cdot v}$, thus $\widetilde{B}_l^* - \widetilde{\overline{B}} = 0$. Therefore, for every $u > v$, i.e., $u = n \cdot v$, the extended distance measure (EDM), $EDM(\widetilde{B}_l^*, \widetilde{\overline{B}})_u = \left[\frac{1}{n\cdot v}\sum_{i=1}^{n\cdot v}[b_{il}^* - (1-b_i)]^2\right]^{1/2} = 0$. Then the extended vectorial distance measure $\widetilde{C}_l = EDM(\widetilde{B}_l^*, \widetilde{\overline{B}})_u \times P_l = 0$, then the quasi-quasi-approximate reasoning results is obtained as $\widetilde{A}_l^{**} = \widetilde{\overline{A}}_l + \widetilde{C}_l = \widetilde{\overline{A}}_l$, and then the quasi-approximate reasoning results is computed as $\widetilde{A}_l^{**} \to \widetilde{A}_l^*$, i.e., $[\widetilde{a}_{lk}^{**}]_{k\times 1} \to [\widetilde{a}_{lp}^*]_{u\times 1}$. From the individual approximate reasoning result $A_l^* = \frac{\widetilde{A}_l^* - \eta_l}{\xi_l - \eta_l}, l = 1$, the final reasoning result is obtained as $A^* = \overline{A} = [1-a_1, \cdots, 1-a_p, \cdots, 1-a_u]$. Therefore ① of this theorem 3.2 is true.

② When the index $u < v$, i.e., $v = n \cdot u$ then the difference between the extended every vectors $\widetilde{B}^*$ and $\widetilde{\overline{B}}$ are calculated as follows;

$$\widetilde{B}_l^* - \widetilde{\overline{B}} = \left[\frac{b_{1l}^* - (1-b_1)}{y_1}, \frac{b_{2l}^* - (1-b_2)}{y_2}, \cdots, \frac{b_{jl}^* - (1-b_j)}{y_j}, \cdots, \frac{b_{v-1,l}^* - (1-b_{v-1})}{y_{v-1}}, \frac{b_{v,l}^* - (1-a_v)}{y_v}\right]$$, that is,

$\widetilde{B}_l^* - \widetilde{\overline{B}} = [b_{1l}^* - (1-b_1), b_{2l}^* - (1-b_2), \cdots, b_{jl}^* - (1-b_j), \cdots, b_{v-1,l}^* - (1-b_{v-1}), b_{v,l}^* - (1-b_v)]$.

Since $B^*(y) = \overline{B}(y) \subseteq F(Y), y \in Y$, i.e., $b_{jl}^* = 1 - b_j, j = \overline{1, v}$ then $\widetilde{B}_l^* - \widetilde{\overline{B}} = 0$. Therefore, for every $u < v$, i.e., $v = n \cdot u$, the extended distance measure (EDM), $EDM(\widetilde{B}_l^*, \widetilde{\overline{B}})_v = \left[\frac{1}{v}\sum_{j=1}^{v}[b_{jl}^* - (1-b_j)]^2\right]^{1/2} = 0$. Thereby the extended vectorial distance measure $\widetilde{C}_l = EDM(\widetilde{B}_l^*, \widetilde{\overline{B}})_v \times P_l = 0$, the quasi-quasi-approximate reasoning results is obtained as $\widetilde{A}_l^{**} = \widetilde{\overline{A}}_l + \widetilde{C}_l = \widetilde{\overline{A}}_l$, and then the quasi- approximate reasoning results is computed as $\widetilde{A}_l^{**} \to \widetilde{A}_l^*$, i.e., $[\widetilde{a}_{lk}^{**}]_{k\times1} \to [\widetilde{a}_{lp}^*]_{u\times1}$. From the individual approximate reasoning result $A_l^* = \frac{\widetilde{A}_l^* - \eta_l}{\xi_l - \eta_l}, l=1$, the final reasoning result is obtained as $A^* = \{A_l^*\} = \overline{A},\ l=1$. Therefore ② of this theorem 3.2 is right.

③ The difference between the extended every vectors $\widetilde{B}^*$ and $\widetilde{\widetilde{B}}$ according to the index $u \neq v$, are calculated as follows; $\widetilde{B}_l^* - \widetilde{\widetilde{B}} = \left[\frac{b_{1l}^* - (1-b_1)}{y_1}, \frac{b_{2l}^* - (1-b_2)}{y_2}, \cdots, \frac{b_{rl}^* - (1-b_r)}{y_r}, \cdots, \frac{b_{u\cdot v-1,l}^* - (1-b_{u\cdot v-1})}{y_{n\cdot v-1}}, \frac{b_{u\cdot v,l}^* - (1-b_{u\cdot v})}{y_{n\cdot v}}\right]$, for short, $\widetilde{B}_l^* - \widetilde{\widetilde{B}} = [b_{1l}^* - (1-b_1), b_{2l}^* - (1-b_2), \cdots, b_{rl}^* - (1-b_r), \cdots, b_{u\cdot v-1,l}^* - (1-b_{u\cdot v-1}), b_{u\cdot v,l}^* - (1-b_{u\cdot v})]$. From the condition of theorem 1, since $b_{rl}^* = 1 - b_r, r = \overline{1, u\cdot v}$, following equation $\widetilde{B}_l^* - \widetilde{\widetilde{B}} = 0$ is satisfied. Therefore, for every index $u \cdot v$, the extended distance measure (EDM), $EDM(\widetilde{B}_l^*, \widetilde{\widetilde{B}})_{u\cdot v} = \left[\frac{1}{u\cdot v}\sum_{i=1}^{u\cdot v}[b_{il}^* - (1-b_i)]^2\right]^{1/2} = 0$. Then the extended vectorial distance measure $\widetilde{C}_l = EDM(\widetilde{B}_l^*, \widetilde{\widetilde{B}})_{u\cdot v} \times P_l = 0$, then the quasi-quasi-approximate reasoning results is obtained as $\widetilde{\widetilde{A}}_l^{**} = \widetilde{\widetilde{A}}_l + \widetilde{C}_l = \widetilde{\widetilde{A}}_l$, and then the quasi-approximate reasoning results is computed as $\widetilde{A}_l^{**} \to \widetilde{A}_l^*$, i.e., $[\widetilde{a}_{lk}^{**}]_{k\times1} \to [\widetilde{a}_{lp}^*]_{u\times1}$. From the individual approximate reasoning result $A_l^* = \frac{\widetilde{A}_l^* - \eta_l}{\xi_l - \eta_l}, l=1$, the final reasoning result is obtained as $A^* = \overline{A} = [1-a_1, \cdots, 1-a_p, \cdots, 1-a_u]$. Hence ③ of this theorem 3.2 is right.

Therefore from ①, ②, ③, if the given premise discrete fuzzy vector is $B^*(y) = \overline{B}(y) \subseteq F(Y), y \in Y$, and the proposed fuzzy approximate reasoning method, i.e., FMT-EDM is applied in the SISO fuzzy system, then the fuzzy approximate reasoning result $A^*(x) \subseteq F(X), x \in X$ of FMT-EDM is obtained to equal to the consequent fuzzy vector $\overline{A}(x) \subseteq F(X), x \in X$ of the fuzzy rule, thereby when $F(X), F(Y)$ are all the fuzzy subsets on the universe of discourse $X, Y$, respectively, the reductive property of the proposed FMT-EDM is completely satisfied. (Proof End)

**Experiment 3.4.** For the SISO fuzzy system with discrete fuzzy set vectors of different dimensions, let us consider an approximate fuzzy reasoning based on 7×1 dimension antecedent fuzzy row vector *B*(*y*)=[0, 0, 0, 0, 0, 0.3, 1] and 5×1 dimension consequent fuzzy row vector *A*(*x*)=[1, 0.3, 0, 0, 0] for FMT-EDM. (Table 5)

**Table 5.** FMT-EDM Reasoning Results and Reductive Property in Class 1

| FMT-EDM Premise $B^*(y)$ | | FMT-EDM Reasoning Results and Reductive Property | | | |
|---|---|---|---|---|---|
| | | Reasoning Results $A^*$ | | | *RPCF* |
| $B_1^*$ | [1, 1, 1, 1, 1, 0.7, 0] | $A_1^*$ | *P(+1,0,-1) form* | [0, 0.7, 1, 1, 1] | 100.0(%) |
| | | | *P(+1,-1) form* | [0, 0.7, 1, 1, 1] | 100.0(%) |
| $B_2^*$ | [1, 1, 1, 1, 1, 0.91, 0] | $A_2^*$ | *P(+1,0,-1) form* | [0, 0.64, 0.92, 1, 0.92] | 91.30 (%) |
| | | | *P(+1,-1) form* | [0, 0.7, 1, 1, 1] | 95.80 (%) |
| $B_3^*$ | [1, 1, 1, 1, 1, 0.45, 0] | $A_3^*$ | *P(+1,0,-1) form* | [0, 0.7, 1, 0.9, 1] | 92.98 (%) |
| | | | *P(+1,-1) form* | [0, 0.7, 1, 0.79, 1] | 90.92 (%) |
| $B_4^*$ | [0, 0, 0, 0, 0, 0.3, 1] | $A_4^*$ | *P(+1,0,-1) form* | [0.55, 0.16, 0, 0, 1] | 68.30 (%) |
| | | | *P(+1,-1) form* | [0.55, 0.16, 0, 0, 1] | 68.30 (%) |
| *RPCF* | | *FMT-EDM-P(+1,0,-1) form* | | | 88.15 (%) |
| | | *FMT-EDM-P(+1,-1) form* | | | 88.75 (%) |

**Experiment 3.5.** For the SISO fuzzy system with discrete fuzzy set vectors of different dimensions in Class 2, let us consider an approximate fuzzy reasoning based on 7×1 dimension antecedent fuzzy row vector *B*(*y*)=[0, 0, 0, 0, 0, 0.3, 1] and 5×1 dimension consequent fuzzy row vector *A*(*x*)=[1, 0.3, 0, 0, 0] for FMT-EDM. (Table 6)

**Table 6.** FMT-EDM Reasoning Results and Reductive Property in Class 2

| FMT-EDM Premise $B^*(y)$ | | FMT-EDM Reasoning Results and Reductive Property | | | |
|---|---|---|---|---|---|
| | | Reasoning Results $A^*$ | | | *RPCF* |
| $B_1^*$ | [1, 1, 1, 1, 1, 0.7, 0] | $A_1^*$ | *P(+1,0,-1) form* | [0, 0.7, 1, 1, 1] | 100 % |
| | | | *P(+1,-1) form* | [0, 0.7, 1, 1, 1] | 100 % |
| $B_2^*$ | [1, 1, 1, 1, 1, 0.91, 0] | $A_2^*$ | *P(+1,0,-1) form* | [0, 0.64, 0.92, 1, 0.92] | 91.30 % |
| | | | *P(+1,-1) form* | [0, 0.7, 1, 1, 1] | 95.8 % |
| $B_3^*$ | [1, 1, 1, 1, 1, 0.45, 0] | $A_3^*$ | *P(+1,0,-1) form* | [0, 0.7, 1, 0.9, 1] | 92.98 % |
| | | | *P(+1,-1) form* | [0, 0.7, 1, 0.79, 1] | 90.92 % |
| $B_4^*$ | [0, 0, 0, 0, 0, 0.2, 1] | $A_4^*$ | *P(+1,0,-1) form* | [0.55, 0.11, 0, 0, 1] | 69.13 % |

| | | | |
|---|---|---|---|
| | *P(+1,-1) form* | [0.55, 0.11, 0, 0, 1] | 69.13 % |
| *RPCF* | *FMT-EDM-P(+1,0,-1) form* | | 88.35% |
| | *FMT-EDM-P(+1,-1) form* | | 88.96 % |

**Experiment 3.6.** For the SISO fuzzy system with discrete fuzzy set vectors of different dimensions in Class 1 and Class 2, let us consider an approximate fuzzy reasoning based on the comprehensive reductive property of our proposed method FMT-EDM.(Table 7)

**Table 7.** Comprehension of our proposed FMT-EDM in Class 1 and Class 2

| | *FMT-EDM-P(+1,0,-1)form* | *FMT-EDM-P(+1,-1)form* |
|---|---|---|
| Class 1 | 88.15% | 88.75 % |
| Class 2 | 88.35% | 88.96 % |
| RPCF-average | 88.25% | 88.86% |

**Experiment 3.7.** For the SISO fuzzy system with discrete fuzzy set vectors with different dimensions in Class 1 and Class 2, let us consider the comprehensive reductive property of our proposed EDM method.( Table 8)

**Table 8.** Comprehension of our proposed EDM in Class 1 and Class 2

| | *EDM-P(+1,0,-1)form* | *EDM-P(+1,-1)form* |
|---|---|---|
| RPCF-average | 89.815 % | 90.310% |

From the examples of Table 2 to Table 8, we can know that the reductive property of the proposed EDM method in the SISO fuzzy system with discrete fuzzy set vectors of different dimensions between the antecedent and consequent of the fuzzy rule, about the comprehension of Class 1 and Class 2, is 90.063% for *EDM-P(+1,0,-1)form* and *EDM-P(+1,-1)form*.

## 4. Comparisons of CRI, TIP, QIP, AARS, and Proposed EDM

In this section we check the reductive property of CRI, TIP, QIP, AARS, and the proposed EDM in the SISO fuzzy system with discrete fuzzy set vectors of different dimensions between the antecedent and consequent of the fuzzy rule, in Class 1 and Class 2.

### 4.1. Checking of FMP-QIP and FMT-QIP

In this subsection, we check the reductive property of FMP-QIP and FMT-QIP by applying of the implication of Łukasiewicz, Gödel, $R_0$ and Gougen. It is shown in Table 9.

**Table 9.** Reductive Property of FMP-QIP and FMT-QIP in Class 1

| FMP-QIP Premise $A^*(x)$ | FMP-QIP Reasoning Results $B^*(y)$ and Reductive Property | | | |
|---|---|---|---|---|
| | FMP-QIP-Łukasiewicz | | FMP-QIP-Gödel | |
| [1, 0.3, 0, 0, 0] | [0, 0, 0, 0, 0.3, 1] | 100% | [0, 0, 0, 0, 0.3, 1] | 100% |
| [1, 0.09, 0, 0, 0] | [0, 0, 0, 0, 0.3, 1] | 97% | [0, 0, 0, 0, 0.3, 1] | 97% |
| [1, 0.55, 0, 0, 0] | [0, 0, 0, 0, 0.3, 1] | 96.46% | [0, 0, 0, 0, 0.3, 1] | 96.46% |
| [0, 0.7, 1, 1, 1] | [0, 0, 0, 0, 0.3, 0.5] | 15.71% | [0, 0, 0, 0, 0.3, 0.5] | 15.71% |
| *RPCF-FMP* | 77.29% | | 77.29% | |
| FMT- QIP Premise $B^*(y)$ | FMT- QIP Reasoning Results $A^*(x)$ and Reductive Property | | | |
| | FMT-QIP-Łukasiewicz | | FMT- QIP –Gödel | |
| [1, 1, 1, 1, 1, 0.7, 0] | [0.44, 0.3, 0, 0, 0] | 23.2% | [0.44, 0.3, 0, 0, 0] | 23.20% |
| [1, 1, 1, 1, 1, 0.91, 0] | [0.58, 0.3, 0, 0, 0] | 16.2% | [0.58, 0.3, 0, 0, 0] | 16.20% |
| [1, 1, 1, 1, 1, 0.45, 0] | [0.34, 0.3, 0, 0, 0] | 30.22% | [0.34, 0.3, 0, 0, 0] | 30.22% |
| [0, 0, 0, 0, 0, 0.3, 1] | [1, 0.3, 0, 0, 0] | 100% | [1, 0.3, 0, 0, 0] | 100% |
| *RPCF-FMT* | 42.41% | | 42.41% | |
| FMP-QIP Premise $A^*(x)$ | FMP-QIP Reasoning Results $B^*(y)$ and Reductive Property | | | |
| | FMP-QIP –$R_0$ | | FMP-QIP –Gougen | |
| [1, 0.3, 0, 0, 0] | [0, 0, 0, 0, 0.3, 1] | 100% | [0, 0, 0, 0, 0.3, 1] | 100% |
| [1, 0.09, 0, 0, 0] | [0, 0, 0, 0, 0.3, 1] | 97% | [0, 0, 0, 0, 0.3, 1] | 97% |
| [1, 0.55, 0, 0, 0] | [0, 0, 0, 0, 0.3, 1] | 96.46% | [0, 0, 0, 0, 0.3, 1] | 96.46% |
| [0, 0.7, 1, 1, 1] | [0, 0, 0, 0, 0, 0.5] | 11.43% | [0, 0, 0, 0, 0.3, 0.5] | 15.71% |
| *RPCF FMP* | 76.22% | | 77.29% | |
| FMT- QIP Premise $B^*(y)$ | FMT- QIP Reasoning Results $A^*(x)$ and Reductive Property | | | |
| | FMT- QIP –$R_0$ | | FMT- QIP –Gougen | |
| [1, 1, 1, 1, 1, 0.7, 0] | [0.44, 0.3, 0, 0, 0] | 23.2 % | [0.44, 0.3, 0, 0, 0] | 23.20% |
| [1, 1, 1, 1, 1, 0.91, 0] | [0.58, 0.3, 0, 0, 0] | 16.2% | [0.58, 0.3, 0, 0, 0] | 16.20% |
| [1, 1, 1, 1, 1, 0.45, 0] | [0.3, 0, 0, 0, 0] | 24.95% | [0.34, 0.3, 0, 0, 0] | 30.22% |
| [0, 0, 0, 0, 0, 0.3, 1] | [1, 0.3, 0, 0, 0] | 100% | [1, 0.3, 0, 0, 0] | 100% |
| *RPCF-FMT* | 41.09% | | 42.41% | |

### 4.2. Checking of FMP and FMT by Zadeh's CRI

In this subsection we check the reductive property of FMP and FMT by Zadeh's CRI, i.e., FMP-CRI, FMT-CRI reductive property in Class 1, which is shown in Table 10.

**Table 10.** FMP-CRI, FMT-CRI Reductive Property In Class 1

| FMP-CRI Premise $A^*(x)$ | FMP-CRI Reasoning Results $B^*(y)$ and Reductive Property | | | |
|---|---|---|---|---|
| | FMP- CRI –Łukasiewicz | | FMP- CRI –Gödel | |
| [1, 0.3, 0, 0, 0] | [0, 0, 0, 0, 0, 0.3, 1] | 100% | [0, 0, 0, 0, 0, 0.3, 1] | 100% |
| [1, 0.09, 0, 0, 0] | [0, 0, 0, 0, 0, 0.3, 1] | 97% | [0, 0, 0, 0, 0, 0.3, 1] | 97.00% |
| [1, 0.55, 0, 0, 0] | [0.25, 0.25, 0.25, 0.25, 0.25, 0.55, 1] | 82.15% | [0, 0, 0, 0, 0, 0.51, 1] | 99.42% |
| [0, 0.7, 1, 1, 1] | [1, 1, 1, 1, 1, 1, 1] | 81.43% | [1, 1, 1, 1, 1, 1, 1] | 81.43% |
| *RPCF-FMP* | 90.14% | | 94.46% | |
| FMT-CRI Premise $B^*(y)$ | FMT-CRI Reasoning Results $A^*(x)$ and Reductive Property | | | |
| | FMT-CRI-Łukasiewicz | | FMT-CRI-Gödel | |
| [1, 1, 1, 1, 1, 0.7, 0] | [1, 1, 1, 1, 1] | 74.00% | [1, 1, 1, 1, 1] | 74.00% |
| [1, 1, 1, 1, 1, 0.91, 0] | [1, 1, 1, 1, 1] | 78.20% | [1, 1, 1, 1, 1] | 78.20% |
| [1, 1, 1, 1, 1, 0.45, 0] | [1, 1, 1, 1, 1] | 69.05% | [1, 1, 1, 1, 1] | 69.05% |
| [0, 0, 0, 0, 0, 0.3, 1] | [1, 1, 1, 1, 1] | 24.00% | [1, 1, 1, 1, 1] | 24.00% |
| *RPCF-FMT* | 61.31% | | 61.31% | |
| FMP-CRI Premise $A^*(x)$ | FMP-CRI Reasoning Results $B^*(y)$ and Reductive Property | | | |
| | FMP- CRI –$R_0$ | | FMP- CRI –Gougen | |
| [1, 0.3, 0, 0, 0] | [0, 0, 0, 0, 0, 0.3, 1] | 100% | [0, 0, 0, 0, 0, 0.3, 1] | 100% |
| [1, 0.09, 0, 0, 0] | [0, 0, 0, 0, 0, 0.3, 1] | 97.00% | [0, 0, 0, 0, 0, 0.3, 1] | 97% |
| [1, 0.55, 0, 0, 0] | [0.6, 0.6, 0.6, 0.6, 0.6, 0.6, 1] | 56.40% | [0, 0, 0, 0, 0, 0.55, 1] | 100% |
| [0, 0.7, 1, 1, 1] | [1, 1, 1, 1, 1, 1, 1] | 81.43% | [1, 1, 1, 1, 1, 1, 1] | 81.43% |
| *RPCF-FMP* | 83.71% | | 94.61% | |
| FMT-CRI Premise $B^*(y)$ | FMT-CRI Reasoning Results $A^*(x)$ and Reductive Property | | | |
| | FMT-CRI-$R_0$ | | FMT-CRI-Gougen | |
| [1, 1, 1, 1, 1, 0.7, 0] | [1, 1, 1, 1, 1] | 74.00% | [1, 1, 1, 1, 1] | 74.00% |
| [1, 1, 1, 1, 1, 0.91, 0] | [1, 1, 1, 1, 1] | 78.20% | [1, 1, 1, 1, 1] | 78.20% |
| [1, 1, 1, 1, 1, 0.45, 0] | [1, 1, 1, 1, 1] | 69.05% | [1, 1, 1, 1, 1] | 69.05% |
| [0, 0, 0, 0, 0, 0.3, 1] | [1, 1, 1, 1, 1] | 24.00% | [1, 1, 1, 1, 1] | 24.00% |
| *RPCF-FMT* | 61.31% | | 61.31% | |

### 4.3. Reductive Property Checking of FMP and FMT by Wang's TIP

In Table 11 we show the reductive property of FMP-TIP and FMT-TIP, respectively.

**Table 11.** In Class 1, FMP-TIP, FMT-TIP Reductive Property

| FMP-TIP | FMP- TIP Reasoning Results $B^*(y)$ and Reductive Property | | | |
|---|---|---|---|---|
| Premise $A^*(x)$ | FMP- TIP –Łukasiewicz | | FMP- TIP –Gödel | |
| [1, 0.3, 0, 0, 0] | [0, 0, 0, 0, 0, 0.3, 1] | 100% | [0, 0, 0, 0, 0, 0.3, 1] | 100% |
| [1, 0.09, 0, 0, 0] | [0, 0, 0, 0, 0, 0.3, 1] | 97.00% | [0, 0, 0, 0, 0, 0.3, 1] | 97.00% |
| [1, 0.55, 0, 0, 0] | [0.25, 0.25, 0.25, 0.25, 0.25, 0.55, 1] | 82.15% | [0, 0, 0, 0, 0, 0.51, 1] | 99.42% |
| [0, 0.7, 1, 1, 1] | [1, 1, 1, 1, 1, 1, 1] | 81.43% | [1, 1, 1, 1, 1, 1, 1] | 81.43% |
| *RPCF-FMP* | 90.14% | | 94.46% | |
| FMT- TIP Premise $B^*(y)$ | FMT- TIP Reasoning Results $A^*(x)$ and Reductive Property | | | |
| | FMT- TIP –Łukasiewicz | | FMT- TIP –Gödel | |
| [1, 1, 1, 1, 1, 0.7, 0] | [0, 0, 0, 0, 0] | 26% | [0, 0, 0, 0, 0] | 26% |
| [1, 1, 1, 1, 1, 0.91, 0] | [0, 0, 0, 0, 0] | 21.80% | [0, 0, 0, 0, 0] | 21.80% |
| [1, 1, 1, 1, 1, 0.45, 0] | [0, 0, 0, 0, 0] | 30.95% | [0, 0, 0, 0, 0] | 30.95% |
| [0, 0, 0, 0, 0, 0.3, 1] | [1, 0.3, 0, 0, 0] | 100% | [1, 0.44, 0, 0, 0] | 97.2% |
| *RPCF-FMT* | 44.69% | | 43.99% | |
| FMP- TIP | FMP- TIP –Reasoning Results $B^*(y)$and Reductive Property | | | |

| Premise $A^*(x)$ | FMP- TIP –$R_0$ | | FMP- TIP –Gougen | |
|---|---|---|---|---|
| [1, 0.3, 0, 0, 0] | [0, 0, 0, 0, 0, 0.3, 1] | 100% | [0, 0, 0, 0, 0, 0.3, 1] | 100% |
| [1, 0.09, 0, 0, 0] | [0, 0, 0, 0, 0, 0.3, 1] | 97.00% | [0, 0, 0, 0, 0, 0.3, 1] | 97.00% |
| [1, 0.55, 0, 0, 0] | [0.6, 0.6, 0.6, 0.6, 0.6, 0.6, 1] | 56.40% | [0, 0, 0, 0, 0, 0.55, 1] | 100% |
| [0, 0.7, 1, 1, 1] | [1, 1, 1, 1, 1, 1, 1] | 81.43% | [1, 1, 1, 1, 1, 1, 1] | 81.43% |
| *RPCF-FMP* | 83.71% | | 94.61% | |
| FMT- TIP Premise $B^*(y)$ | FMT- TIP Reasoning Results $A^*(x)$ and Reductive Property | | | |
| | FMT- TIP –$R_0$ | | FMT- TIP –Gougen | |
| [1, 1, 1, 1, 1, 0.7, 0] | [0, 0, 0, 0, 0] | 26% | [0, 0, 0, 0, 0] | 26% |
| [1, 1, 1, 1, 1, 0.91, 0] | [0, 0, 0, 0, 0] | 21.80% | [0, 0, 0, 0, 0] | 21.80% |
| [1, 1, 1, 1, 1, 0.45, 0] | [0, 0, 0, 0, 0] | 30.95% | [0, 0, 0, 0, 0] | 30.95% |
| [0, 0, 0, 0, 0, 0.3, 1] | [1, 0.3, 0, 0, 0] | 100% | [1, 0.3, 0, 0, 0] | 100% |
| *RPCF-FMT* | 44.69% | | 44.69% | |

### 4.4. Checking of FMP and FMT by Turksen and Zhong's AARS

In this subsection, we check the reductive property of FMP and FMT by Turksen and Zhong's AARS. (Table 12)

**Table 12.** In Class 1, FMP-AARS and FMT-AARS Reductive Property

| FMP-AARS Premise $A^*(x)$ | FMP-AARS Reasoning Results and Reductive Property | | |
|---|---|---|---|
| | Reasoning Results $B^*(y)$ | | *RPCF* |
| [1, 0.3, 0, 0, 0] | *more or less form* | [0, 0, 0, 0, 0, 0.3, 1] | 100(%) |
| | *reduction form* | [0, 0, 0, 0, 0, 0.3, 1] | 100(%) |
| [1, 0.09, 0, 0, 0] | *more or less form* | [0, 0, 0, 0, 0, 0.33, 1] | 96.54 (%) |
| | *reduction form* | [0, 0, 0, 0, 0, 0.27, 0.9] | 96.03(%) |
| [1, 0.55, 0, 0, 0] | *more or less form* | [0, 0, 0, 0, 0, 0.34, 1] | 96.99 (%) |
| | *reduction form* | [0, 0, 0, 0, 0, 0.27, 0.89] | 94.44 (%) |
| [0, 0.7, 1, 1, 1] | *more or less form* | [0, 0, 0, 0, 0, 0.56, 1] | 12.30 (%) |
| | *reduction form* | [0, 0, 0, 0, 0, 0.16, 0.53] | 13.22 (%) |
| *RPCF-FMP* | *FMP-AARS more or less form* | | 76.46 (%) |
| | *FMP-AARS reduction form* | | 75.92 (%) |
| FMT-AARS Premise $B^*(y)$ | FMT-AARS Reasoning Results and Reductive Property | | |
| | Reasoning Results $A^*(x)$ | | *RPCF* |
| [1, 1, 1, 1, 1, 0.7, 0] | *more or less form* | [1, 0.57, 0, 0, 0] | 17.46% |
| | *reduction form* | [0.52, 0.16, 0, 0, 0] | 18.67% |
| [1, 1, 1, 1, 1, 0.91, 0] | *more or less form* | [1, 0.58, 0, 0, 0] | 13.32% |
| | *reduction form* | [0.52, 0.16, 0, 0, 0] | 14.51% |
| [1, 1, 1, 1, 1, 0.45, 0] | *more or less form* | [1, 0.57, 0, 0, 0] | 17.67% |
| | *reduction form* | [0.53, 0.16, 0, 0, 0] | 23.57% |
| [0, 0, 0, 0, 0, 0.3, 1] | *more or less form* | [1, 0.3, 0, 0, 0] | 100% |
| | *reduction form* | [1, 0.3, 0, 0, 0] | 100% |
| *RPCF-FMT* | *FMT-AARS more or less form* | | 37.11% |
| | *FMT-AARS reduction form* | | 39.19% |

### 4.5. Comprehensive Comparisons of CRI, TIP, QIP, AARS and Proposed EDM in Class 1

The reductive properties of CRI, TIP, QIP, AARS, and proposed EDM in Class 1 are shown in Table 13.

**Table 13**. CRI, TIP, QIP, AARS, and proposed EDM Reductive Properties in Class 1

| No | In Class 1 Reasoning Method | | *RPCF-FMP* | *RPCF-FMT* | *$RPCF_{FR}$* | Average |
|---|---|---|---|---|---|---|
| 1 | Proposed EDM | *P(+1,0,-1) form* | 87.00% | 88.15% | 87.58% | 87.865% |
| 2 | | *P(+1,-1) form* | 87.54 % | 88.75% | 88.15% | |
| 3 | CRI (1975) | Gödel; G | 94.46 % | 61.31 % | 77.89% | 76.023% |
| 4 | | Gougen; Go | 94.61 % | 61.31 % | 77.96% | |
| 5 | | Łukasiewicz; L | 90.14 % | 61.31 % | 75.73% | |
| 6 | | $R_0$ | 83.71 % | 61.31% | 72.51% | |
| 7 | TIP | Gödel; G | 94.46 % | 43.99% | 69.23% | 67.625% |
| 8 | | Gougen; Go | 94.61 % | 44.69 % | 69.65% | |
| 9 | (1999) | Łukasiewicz | 90.14 % | 44.69% | 67.42% | |
| 10 | | $R_0$ | 83.71 % | 44.69% | 64.20% | |
| 11 | QIP (2015-2018) | Łukasiewicz | 77.29% | 42.41% | 59.85% | 59.255% |
| 12 | | Gödel; G | 77.29% | 42.41% | 59.85% | |
| 13 | | $R_0$ | 77.29% | 42.41% | 59.85% | |
| 14 | | Gougen; Go | 76.22 % | 41.09% | 58.66% | |
| 15 | AARS (1990) | *reduction form* | 75.92 % | 39.19 % | 57.56% | 57.175% |
| 16 | | *more or less form* | 76.46 % | 37.11 % | 56.79% | |

Through the Matlab experiments we can obtain the following propositions in Class 1.

**Proposition 4.1.** For FMP, the reductive property of CRI-Gougen and TIP-Gougen among 16 fuzzy approximate reasoning methods is best high in Class 1 for SISO fuzzy system with discrete fuzzy set vectors of different dimensions.

**Proposition 4.2.** For FMP, the reductive property of QIP-Gougen among 16 fuzzy approximate reasoning methods is best low in Class 1 for SISO fuzzy system with discrete fuzzy set vectors of different dimensions.

**Proposition 4.3.** For FMT, the reductive property of EDM-*P(+1,0,-1) form* and EDM-*P(+1,-1) form* among 16 fuzzy approximate reasoning methods is high low in Class 1 for SISO fuzzy system with discrete fuzzy set vectors of different dimensions.

**Proposition 4.4.** For FMT, the reductive property of AARS- *reduction form* and AARS-*more or less form* is best low in Class 1 for SISO fuzzy system with discrete fuzzy set vectors of different dimensions.

**Proposition 4.5.** For FMT and FMT, the reductive property of EDM among 5 fuzzy approximate reasoning methods is best high, and then CRI, TIP, in Class 1 for SISO fuzzy system with discrete fuzzy set vectors of different dimensions.

**Proposition 4.6.** For FMT and FMT, the reductive property of AARS among 5 fuzzy approximate reasoning methods is best low in Class 1 for SISO fuzzy system with discrete fuzzy set vectors of different dimensions.

### 4.6. Comparisons of CRI, TIP, QIP, AARS and Proposed Method in Class 2

In this subsection, we compare and analyze about CRI, TIP, QIP, AARS and proposed EDM method for Class 2. The reductive properties of five fuzzy reasoning methods for Class 2 are shown in Table 14. Computational process is omitted in this paper.

**Table 14.** Reductive Properties of CRI, TIP, QIP, AARS and DMM in Class 2

| No | In Class 1 Reasoning Method | | *RPCF-FMP* | *RPCF-FMT* | $RPCF_{FR}$ | Average |
|---|---|---|---|---|---|---|
| 1 | Proposed EDM | *P(+1,0,-1) form* | 95.76% | 88.35% | 92.060% | 92.265% |
| 2 | | *P(+1,-1) form* | 95.97% | 88.96% | 92.470% | |
| 3 | CRI (1975) | Gödel; G | 98.75% | 61.31% | 80.030% | 78.163% |
| 4 | | Gougen; Go | 98.89% | 61.31% | 80.100% | |
| 5 | | Łukasiewicz; L | 94.43% | 61.31% | 77.870% | |
| 6 | | $R_0$ | 87.99% | 61.31% | 74.650% | |
| 7 | TIP (1999) | Gödel; G | 98.89% | 43.02% | 70.967% | 67.948% |
| 8 | | Gougen; Go | 98.89% | 34.36% | 66.6240% | |
| 9 | | Łukasiewicz | 94.43% | 44.19% | 69.309% | |
| 10 | | $R_0$ | 87.99% | 41.79% | 64.890% | |
| 11 | QIP (2015-2018) | Łukasiewicz | 98.01% | 41.91% | 69.957% | 69.629% |
| 12 | | Gödel; G | 98.01% | 41.91% | 69.957% | |
| 13 | | $R_0$ | 98.01% | 40.59% | 69.300% | |
| 14 | | Gougen; Go | 98.01% | 41.91% | 69.957% | |
| 15 | AARS (1990) | *reduction form* | 74.76% | 39.19% | 56.970% | 56.64% |
| 16 | | *more or less form* | 75.56% | 37.07% | 56.310% | |

From the Matlab experiments we can obtain the following propositions in Class 2.

**Proposition 4.7.** For FMP, the reductive property of CRI-Gougen, TIP-Gougen and TIP-Gödel among 16 fuzzy approximate reasoning methods is best high in Class 2 for SISO fuzzy system with discrete fuzzy set vectors of different dimensions.

**Proposition 4.8.** For FMP, the reductive property of AARS-*reduction form* and AARS-*more or less form* among 16 fuzzy approximate reasoning methods is best low in Class 2 for SISO fuzzy system with discrete fuzzy set vectors of different dimensions.

**Proposition 4.9.** For FMT, the reductive property of EDM-*P(+1,0,-1) form* and EDM-*P(+1,-1) form* among 16 fuzzy approximate reasoning methods is best high in Class 2 for SISO fuzzy system with discrete fuzzy set vectors of different dimensions.

**Proposition 4.10.** For FMT, the reductive property of AARS-*reduction form* and AARS-*more or less form* is best low in Class 2 for SISO fuzzy system with discrete fuzzy set vectors of different dimensions.

**Proposition 4.11.** For FMP and FMT, the reductive property of EDM among 5 fuzzy approximate reasoning methods is best high, and then CRI, TIP, in Class 2 for SISO fuzzy system with discrete fuzzy set vectors of different dimensions.

**Proposition 4.12.** For FMP and FMT, the reductive property of AARS among 5 fuzzy approximate reasoning methods is best low in Class 2 for SISO fuzzy system with discrete fuzzy set vectors of different dimensions.

### 4.7. Comprehensive Analysis of the reductive property in Class 1 and Class 2

The comparison of the fuzzy approximate reasoning computing times is checked for CRI, TIP, QIP, AARS, and proposed our EDM. The experiment was accomplished via $6^{th}$ of test, these average value is shown in Table 15.

**Table 15.** Comparison of the fuzzy approximate reasoning computing times

| | T1(ms) | T2(ms) | T3(ms) | T4(ms) | T5(ms) | T6(ms) | Average(ms) |
|---|---|---|---|---|---|---|---|
| AARS | 222 | 232 | 252 | 225 | 233 | 231 | 233 |
| Proposed EDM | 257 | 267 | 242 | 232 | 260 | 243 | 250 |
| CRI | 307 | 254 | 254 | 239 | 245 | 233 | 255 |
| TIP | 275 | 284 | 262 | 251 | 253 | 237 | 260 |
| QIP | 298 | 278 | 288 | 297 | 266 | 263 | 282 |

**Proposition 4.13.** As known in Table 4.7 AARS's computational time is best shorted, and then our proposed EDM method, CRI, TIP, and QIP's one is longest.

The total reductive properties of the 5 fuzzy reasoning methods in Class 1 and Class 2 are comprehensively shown in **Fig. 1**. From **Fig. 1** we can see the following propositions in Class 1 and Class 2.

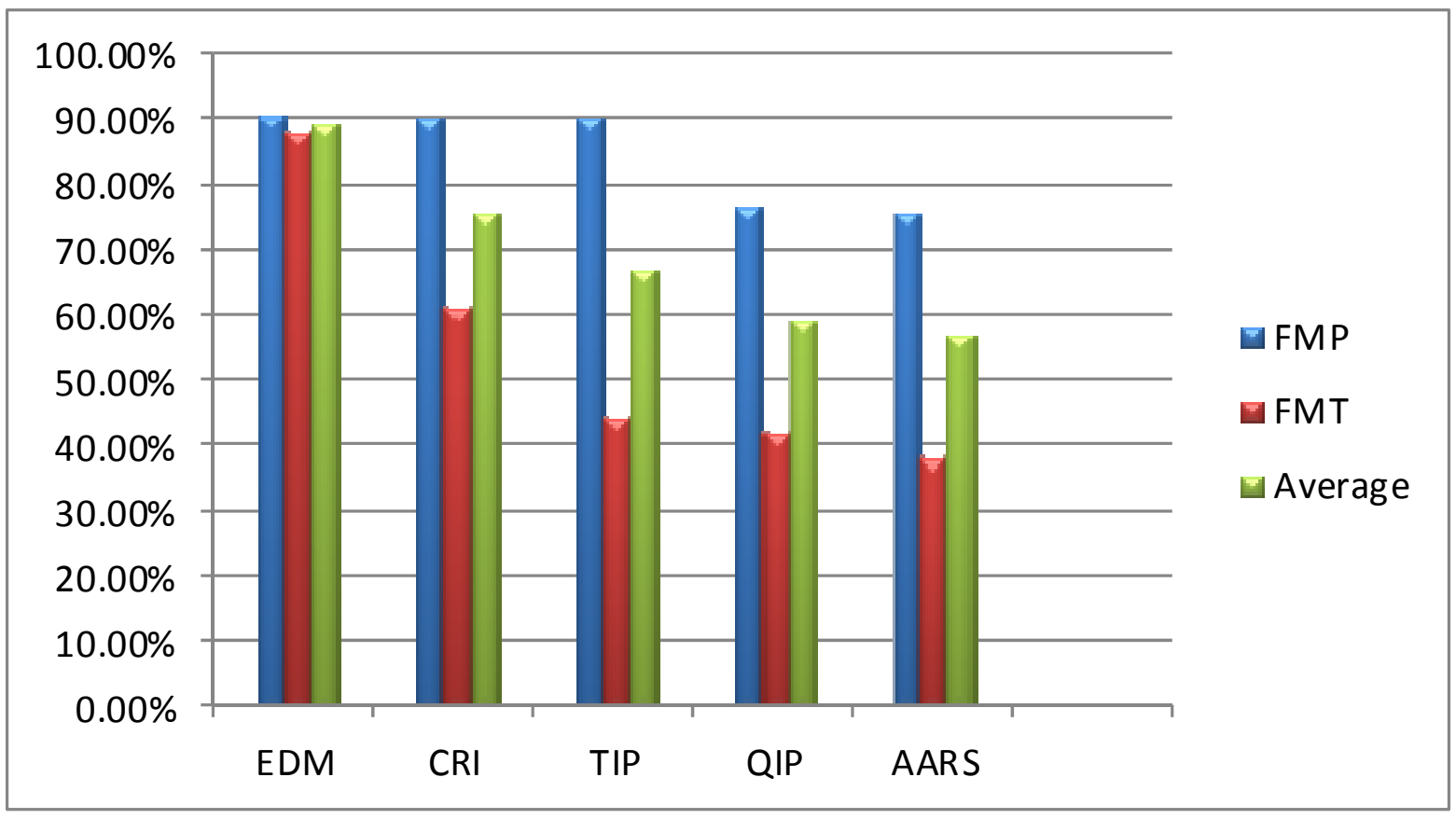

| | FMP | FMT | AVERAGE |
|---|---|---|---|
| Proposed EDM | 91.57% | 88.56% | 90.07% |
| CRI | 92.87% | 61.31% | 77.09% |
| TIP | 92.87% | 42.68% | 67.77% |
| QIP | 81.22% | 41.83% | 61.53% |
| AARS | 75.68% | 38.14% | 56.91% |

**Fig. 1**. The comprehensive reductive properties of the 5 fuzzy reasoning methods for Class 1 and Class 2

**Proposition 4.14.** For FMP, Class 1 and Class 2, the reductive property of CRI and TIP among 5 fuzzy approximate reasoning methods is best high for SISO fuzzy system with discrete fuzzy set vectors of different dimensions.

**Proposition 4.15.** For FMP, Class 1 and Class 2, the reductive property of AARS among 5 fuzzy approximate reasoning methods is best low for SISO fuzzy system with discrete fuzzy set vectors of different dimensions.

**Proposition 4.16.** For FMT, Class 1 and Class 2, the reductive property of EDM among 5 fuzzy approximate reasoning methods is best high for SISO fuzzy system with discrete fuzzy set vectors of different dimensions.

**Proposition 4.17.** For FMT, Class 1 and Class 2, the reductive property of AARS is best low for SISO fuzzy system with discrete fuzzy set vectors of different dimensions.

**Proposition 4.18.** For FMP and FMT, Class 1 and Class 2, among 5 fuzzy approximate reasoning methods, the reductive property of EDM is best high, and then CRI, TIP, for SISO fuzzy system with discrete fuzzy set vectors of different dimensions.

**Proposition 4.19.** For FMP and FMT, Class 1 and Class 2, among 5 fuzzy approximate reasoning methods, the reductive property of AARS is best low for SISO fuzzy system with discrete fuzzy set vectors of different dimensions.

Comprehensively among the 5 fuzzy reasoning methods for Class 1 and Class 2, our proposed EDM method is highest with respect to the reductive property.

# 5. Conclusions

In this paper our research results can be summarized as follows.

Firstly, in this paper we presented a novel original method of fuzzy approximate reasoning that can open a new direction of research in the uncertainty inference of AI and CI, which is based on distance measure, concretely, an extended distance measure(EDM) method, with smaller information loss. And the Distance Measure Method(DMM) presented in the paper [12,13] is a special case of this paper, that is, index of the antecedent and the given premise is $u = v$, i.e., $n = 1$ in the formula (17) and (38).

Secondly, we proposed a novel fuzzy approximate reasoning method based on an extended distance measure in the SISO fuzzy system with discrete fuzzy set vectors of different dimensions between the antecedent and consequent. We call this method an EDM one. That is, EDM based fuzzy approximate reasoning method is consisted of two part, i.e., FMP-EDM, and FMT-EDM.

Thirdly, we proved two theorems about the reductive property of the FMP-EDM, and FMT-EDM, and then demonstrated several examples. In this paper, discrete fuzzy set vectors of different dimensions mean that the dimension of the antecedent discrete fuzzy set differs from one of the consequent discrete fuzzy set in the SISO fuzzy system.

Fourthly, we compared the reductive properties for 5 fuzzy reasoning methods with respect to FMP and FMT, which are CRI, TIP, QIP, AARS, and an our proposed EDM. Through the illustrative experiments by Matlab, we obtained the conclusion that our proposed method EDM have the highest reductive property, the smallest information loss, and comparatively small computational time. The experimental results highlight that the proposed approximate reasoning method EDM is comparatively clear and effective, and in accordance with human thinking than the existing fuzzy reasoning methods.

# Appendix

Matlab program for the realization of our proposed EDM method is as follows.

```
clc
clear all

A = [1 0.3 0 0 0]; % 5*1 row vector
B = [0 0 0 0 0 0.3 1]; % 7*1 row vector

size_A = size(A,2); %The size of the fuzzy set A
size_B = size(B,2); %The size of the fuzzy set B
Fact_A = ExtendFactor2(size_A, size_B); % calculate the extend factor of the fuzzy set A.

Fact_B = Fact_A * size_A / size_B; % calculate the extend factor of the fuzzy set B.

%---Calculate the extended fuzzy set of A,B-------
Ad = [];
Bd = [];
Ad = Ext2_vector(A , size_A , Fact_A); % Calculate the 35*1 extended fuzzy set of A
Bd = Ext2_vector(B , size_B , Fact_B); % Calculate the 35*1 extended fuzzy set of B


A1 =  [1, 0.2, 0, 0, 0]; % the slightly tilted set of A
B1 =[0, 0, 0, 0, 0, 0.2, 1]; % the slightly tilted set of B

%---- Calculate the extended fuzzy set of the s.t.A and st.B------------
A1d = [];
B1d=[];
A1d = Ext2_vector(A1 , size_A , Fact_A);
B1d = Ext2_vector(B1 , size_B , Fact_B);
AA = [1-A; 1-A.^2; 1-sqrt(A); A];
BB = [B; B.^2; sqrt(B); 1-B];


%  -----Tilted---small---large--- FMP EDM class 1---------------
```

```
Adstar = [Ad; Ad.^2; sqrt(Ad); 1-Ad]; % The extended fuzzy set of the given premise for FMP-EDM;case
1, 2, 3, 4;
Astar = [A; A.^2; sqrt(A); 1-A]; % The given premise for FMP-EDM
BBd = [Bd; Bd.^2; sqrt(Bd); 1-Bd]; %The extended fuzzy set of the consequent for FMP-EDM

n = size(Adstar, 1); % The number of the row of the extended fuzzy set of the given premise
CF_fmp = zeros(1,n);

for i = 1:n
   EDM = rms(Adstar(i,:)-Ad); % Calculate the extended distance measure.
   P = sign(Adstar(i,:)-Ad); % Calculate the sign vector.
   % compute the quasi-quasi-approximate reasoning results
   if i==4
      beta = 1 - Bd + EDM*P; % if case==4
   else
      beta = Bd + EDM*P; % if case==1 or 2 or 3
   end
   % end the calculation of the quasi-quasi-approximate reasoning results
   beta = re2_vector(beta , 1 , size_B , Fact_B); % the quasi-approximate reasoning results
   B1star(i,:) = standard_vector(beta); %the individual approximate reasoning result
   CF_fmp(i) = RPCF(B1star(i,:), BB(i,:)); % Calculate the evaluation for the reductive property
end

infResult('FMP EDM class1', A1,B, Astar,B1star, CF_fmp);

%  ------Tilted---small---large---- FMT EDM class 1 -------------

B1dstar = [1-Bd; 1-Bd.^2; 1-sqrt(Bd); Bd]; % The extended fuzzy set of the given premise for FMT-EDM
Bstar = [1 - B ; 1 - B.^2 ; 1 - sqrt(B) ; B]; % The given premise for FMT-EDM
AAd = [1-Ad; 1-Ad.^2; 1-sqrt(Ad); Ad]; %The extended fuzzy set of the consequent for FMT-EDM
AA = [1-A; 1-A.^2; 1-sqrt(A); A]; %The consequent for FMT-EDM

Adstar = zeros(size(AAd)); %Formate the quasi-quasi-approximate reasoning results
CF_fmt = zeros(1,n);

for i = 1:n
   EDM = rms(B1dstar(i,:)-(1-Bd)); % Calculate the extended distance measure.
   P = sign(B1dstar(i,:)-(1-Bd)); % Calculate the sign vector.
  % compute the quasi-quasi-approximate reasoning results
  if i==4
      alpha = 1-Ad + EDM*P; % if case=9
   else
      alpha = 1-Ad + EDM*P; % if case=6 or 7 or 8
  end
    % end the calculation of the quasi-quasi-approximate reasoning results
   alpha = re2_vector(alpha , 1 , size_A , Fact_A); % the quasi-approximate reasoning results
   Astar(i,:) = standard_vector(alpha); %the individual approximate reasoning result
   CF_fmt(i) = RPCF(Astar(i,:), AA(i,:)); % Calculate the evaluation for the reductive property
end


infResult('FMT EDM class1', A , B, Astar,Bstar, CF_fmt);

disp(['Total: ', num2str(mean([mean(CF_fmp) mean(CF_fmt)])), ' %']);


%  ---------tilted-----small---large------ FMP EDM class 2 -----KSI-------

Adstar = [Ad; Ad.^2; sqrt(Ad); A1d];% The extended fuzzy set of the given premise for FMP-EDM ; case
1, 2, 3, 5
Astar = [A; A.^2; sqrt(A); A1]; % The given premise for FMP-EDM
BBd = [Bd; Bd.^2; sqrt(Bd); B1d]; %The extended fuzzy set of the consequent for FMP-EDM
BB = [B; B.^2; sqrt(B); B1]; %The consequent for FMP-EDM

CF_fmp = zeros(1,n);
for i = 1:n
   EDM = rms(Adstar(i,:)-Ad); % Calculate the extended distance measure.
   P = sign(Adstar(i,:)-Ad); % Calculate the sign vector.
   % compute the quasi-quasi-approximate reasoning results
  if i==4
      beta = B1d + EDM*P; % if case==5
   else
      beta = Bd + EDM*P; % if case==1 or 2 or 3
   end
    % end the calculation of the quasi-quasi-approximate reasoning results
   beta = re2_vector(beta , 1 , size_B , Fact_B); % the quasi-approximate reasoning results
   Bstar(i,:) = standard_vector(beta); %the individual approximate reasoning result
```

```
    CF_fmp(i) = RPCF(Bstar(i,:), BB(i,:)); % Calculate the evaluation for the reductive property
end

infResult('FMP EDM class2', A, B, Astar, Bstar, CF_fmp);


% ------tilted---small---large---- FMT EDM class 2 ---KSI------
Bdstar = [1-Bd; 1-Bd.^2; 1-sqrt(Bd); B1d];% The extended fuzzy set of the given premise for FMP-EDM ;
case 6, 7, 8, 10
Bstar = [1-B; 1-B.^2; 1-sqrt(B); B1];% The given premise for FMT-EDM
AAd = [1-Ad; 1-Ad.^2; 1-sqrt(Ad); A1d];%The extended fuzzy set of the consequent for FMT-EDM
AA = [1-A; 1-A.^2; 1-sqrt(A); A1]; %The consequent for FMP-EDM

Adstar = zeros(size(Bdstar));
CF_fmt = zeros(1,n);

for i = 1:n
    EDM = rms(Bdstar(i,:)-(1-Bd)); % Calculate the extended distance measure.
    P = sign(Bdstar(i,:)-(1-Bd)); % Calculate the sign vector.
    % compute the quasi-quasi-approximate reasoning results
    if i==4
        alpha = A1d + EDM*P; % if case==10
    else
        alpha = 1-Ad + EDM*P; % if case==6 or 7 or 8
    end
     % end the calculation of the quasi-quasi-approximate reasoning results
    alpha = re2_vector(alpha , 1 , size_A , Fact_A); % the quasi-approximate reasoning results
    Astar(i,:) = standard_vector(alpha); %the individual approximate reasoning result
    CF_fmt(i) = RPCF(Astar(i,:), AA(i,:)); % Calculatethe evaluation for the reductive property
end

infResult('FMT EDM class2', A , B, Astar,Bstar, CF_fmt);

disp(['Total: ', num2str(mean([mean(CF_fmp) mean(CF_fmt)])), ' %']);
```

```
=====================================================================
FMP EDM class1
   A* = A    = [1, 0.3, 0, 0, 0] --> B*=[0, 0, 0, 0, 0, 0.3, 1]    100 %
   A*= A^2 = [1, 0.09, 0, 0, 0] --> B*=[0.097, 0, 0, 0, 0.097, 0.37, 1]    93.25 %
   A*=A^0.5= [1, 0.55, 0, 0, 0] --> B*=[0, 0.12, 0.12, 0.12, 0, 0.3, 1]    91.21 %
   A* ~    A = [0, 0.7, 1, 1, 1] --> B*=[0, 0, 1, 1, 1, 0.83, 0.43]    63.53 %
RPCF-FMP EDM class1---sign -average: 87 %
+--------------------------------------------------------------------------+
FMT EDM class1
   B* =    1-B = [1, 1, 1, 1, 1, 0.7, 0] --> A*=[0, 0.7, 1, 1, 1]    100 %
   B*= 1-B^2 = [1, 1, 1, 1, 1, 0.91, 0] --> A*=[0, 0.64, 0.92, 1, 0.92]    91.3 %
   B*=1-B^0.5= [1, 1, 1, 1, 1, 0.45, 0] --> A*=[0, 0.7, 1, 0.9, 1]    92.98 %
   B*    ~    B    = [0, 0, 0, 0, 0, 0.3, 1] --> A*=[0, 0.25, 0.35, 0.35, 1]    44.78 %
RPCF-FMT EDM class1---sign -average: 82.27 %
+--------------------------------------------------------------------------+
Total: 84.6331 %
FMP EDM class2
   A* = A    = [1, 0.3, 0, 0, 0] --> B*=[0, 0, 0, 0, 0, 0.3, 1]    100 %
   A*= A^2 = [1, 0.09, 0, 0, 0] --> B*=[0.097, 0, 0, 0, 0.097, 0.37, 1]    93.25 %
   A*=A^0.5= [1, 0.55, 0, 0, 0] --> B*=[0, 0.12, 0.12, 0.12, 0, 0.3, 1]    91.21 %
   A* ~    A = [1, 0.2, 0, 0, 0] --> B*=[0.035, 0, 0, 0, 0.035, 0.23, 1]    98.58 %
RPCF-FMP EDM class2---sign -average: 95.76 %
+--------------------------------------------------------------------------+
FMT EDM class2
   B* =    1-B = [1, 1, 1, 1, 1, 0.7, 0] --> A*=[0, 0.7, 1, 1, 1]    100 %
   B*= 1-B^2 = [1, 1, 1, 1, 1, 0.91, 0] --> A*=[0, 0.64, 0.92, 1, 0.92]    91.3 %
   B*=1-B^0.5= [1, 1, 1, 1, 1, 0.45, 0] --> A*=[0, 0.7, 1, 0.9, 1]    92.98 %
   B*    ~    B    = [0, 0, 0, 0, 0, 0.2, 1] --> A*=[0.55, 0.11, 0, 0, 1]    69.13 %
RPCF-FMT EDM class2---sign -average: 88.35 %
+--------------------------------------------------------------------------+
Total: 92.058 %
>>
```